\documentclass[journal]{IEEEtran}

\usepackage[utf8]{inputenc}
\usepackage[T1]{fontenc}
\usepackage{microtype}
\usepackage[dvips]{graphicx}
\usepackage{xcolor}
\usepackage{times}
\usepackage{float}
\usepackage{amsmath}
\usepackage{amssymb}
\usepackage{mathabx}
\usepackage{dsfont}
\usepackage[ruled,vlined,noresetcount]{algorithm2e}
\usepackage{cancel}
\usepackage{multirow}
\usepackage{subcaption}
\usepackage{blindtext}
\usepackage[colorlinks = true,
            linkcolor = blue,
            urlcolor  = blue,
            citecolor = blue,
            anchorcolor = blue]{hyperref}
\usepackage{amssymb}
\usepackage{tikz}
\usetikzlibrary{shapes,arrows,calc,positioning}
\usepackage{makecell}
\usepackage{multirow}
\usepackage{array}
\usepackage{nicefrac}
\newcommand{\PreserveBackslash}[1]{\let\temp=\\#1\let\\=\temp}
\newcolumntype{C}[1]{>{\PreserveBackslash\centering}p{#1}}
\newcolumntype{R}[1]{>{\PreserveBackslash\raggedleft}p{#1}}
\newcolumntype{L}[1]{>{\PreserveBackslash\raggedright}p{#1}}

\ifCLASSINFOpdf
\else
\fi

\usepackage{algpseudocode}

\begin{document}
%

\title{Model-Based Policy Search Using Monte Carlo Gradient Estimation with Real Systems Application}

\author{Fabio Amadio$^1$, Alberto Dalla Libera$^1$, Riccardo Antonello$^1$, Daniel Nikovski$^2$, Ruggero Carli$^1$, Diego Romeres$^2$
    \thanks{$^1$ Fabio Amadio, Alberto Dalla Libera, Riccardo Antonello and Ruggero Carli are with the Deptartment of Information Engineering, University of Padova, Via Gradenigo 6/B, 35131 Padova, Italy [fabio.amadio@phd.unipd.it, dallaliber@dei.unipd.it, antonello@dei.unipd.it, carlirug@dei.unipd.it].}%
    \thanks{$^2$ Diego Romeres and Daniel Nikovski are with Mitsubishi Electric Research Laboratories (MERL), Cambridge, MA 02139 [romeres@merl.com, nikovski@merl.com].}%
}%



%


\maketitle

\begin{abstract}
In this paper, we present a Model-Based Reinforcement Learning (MBRL) algorithm named \emph{Monte Carlo Probabilistic Inference for Learning COntrol} (MC-PILCO). The algorithm relies on Gaussian Processes (GPs) to model the system dynamics and on a Monte Carlo approach to estimate the policy gradient. This defines a framework in which we ablate the choice of the following components: (i) the selection of the cost function, (ii) the optimization of policies using dropout, (iii) an improved data efficiency through the use of structured kernels in the GP models. The combination of the aforementioned aspects affects dramatically the performance of MC-PILCO. Numerical comparisons in a simulated cart-pole environment show that MC-PILCO exhibits better data efficiency and control performance w.r.t. state-of-the-art GP-based MBRL algorithms. Finally, we apply MC-PILCO to real systems, considering in particular systems with partially measurable states. We discuss the importance of modeling both the measurement system and the state estimators during policy optimization. The effectiveness of the proposed solutions has been tested in simulation and on two real systems, a Furuta pendulum and a ball-and-plate rig. MC-PILCO code is publicly available at \url{https://www.merl.com/research/license/MC-PILCO}.
\end{abstract}

%
\IEEEpeerreviewmaketitle

\section{Introduction}\label{sec:introduction}
In recent years, reinforcement learning (RL) \cite{sutton2018reinforcement} has achieved outstanding results in many different environments, and has shown the potential to provide an automated framework for learning different controllers by self-experimentation. However, model-free RL (MFRL) algorithms might require a massive amount of interactions with the environment in order to solve the assigned task. This data inefficiency puts a limit to RL's potential in real-world applications, due to the time and cost of interacting with them. In particular, when dealing with mechanical systems, it is critical to learn the task with the least possible amount of interaction, to reduce wear and tear and avoid any damage to the system. A promising way to overcome this limitation is model-based reinforcement learning (MBRL). MBRL is based on the use of data from interactions to build a predictive model of the environment and to exploit it to plan control actions. MBRL increases data efficiency by using the model to extract more valuable information from the available data \cite{atkeson1997comparison}.

On the other hand, MBRL methods are effective only inasmuch as their models resemble accurately the real systems. Hence, deterministic models might suffer dramatically from model inaccuracy, and the use of stochastic models becomes necessary in order to capture uncertainty. Gaussian Processes (GPs) \cite{williams2006gaussian} are a class of Bayesian models commonly used in RL methods precisely for their intrinsic capability to handle uncertainty and provide principled stochastic predictions \cite{kuss2004gaussian}\cite{berkenkamp2017safe}.\\

\textbf{Related work.} PILCO (Probabilistic Inference for Learning COntrol) \cite{deisenroth2011pilco} is a successful MBRL algorithm that uses GP models and gradient-based policy search to achieve substantial data efficiency in solving different control problems, both in simulation as well as with real systems \cite{deisenroth2011learning}\cite{deisenroth2012toward}. In PILCO, long-term predictions are computed analytically, approximating the distribution of the next state at each time instant with a Gaussian distribution by means of moment matching. In this way, the policy gradient is computed in closed form. However, the use of moment matching introduces two relevant limitations. (i) Moment matching models only unimodal distributions. (ii) The computation of the moments is shown to be tractable only when considering Squared Exponential (SE) kernels and differentiable cost functions. The unimodal approximation is too crude of an assumption on the long-term system dynamics for several systems. Moreover, it introduces relevant limitations in case that initial conditions or the optimal solution are multimodal. For instance, in case that the initial variance of the state distribution is high, the optimal solution might be multimodal, due to dependencies on initial conditions.  Also the limitation on the kernel choice might be very stringent, as the SE kernel imposes smooth properties on the GPs posterior estimator and might show poor generalization properties in data that have not been seen during training \cite{GIP, romeres2019semiparametrical,romeres2016online,SP_peters}.

PILCO has inspired several other MBRL algorithms that try to improve it in different ways. Limitations due to the use of SE kernels have been addressed in Deep-PILCO \cite{gal2016improving}, where the system evolution is modeled using Bayesian Neural Networks \cite{mackay1992bayesian}, and long-term predictions are computed combining particle-based methods and moment matching. Results show that, compared to PILCO, Deep-PILCO requires a larger number of interactions with the system in order to learn the task. This fact suggests that using neural networks (NNs) might not be advantageous in terms of data efficiency, due to the considerably high amount of parameters needed to characterize the model. A more articulated approach has been proposed in PETS \cite{chua2018deep}, where the authors use a probabilistic ensemble of NNs to model the uncertainty of the system dynamics. Despite the positive results in the simulated high-dimension systems, also the numerical results in PETS show that GPs are more data-efficient than NNs when considering low-dimensional systems, such as the cart-pole benchmark. An alternative route has been proposed in \cite{cutler2015efficient}, where the authors use a simulator to learn a prior for the GP model before starting the reinforcement learning procedure on the actual system to control. This simulated prior improves the performance of PILCO in areas of the state space with no available data points. However, the method requires an accurate simulator that may not always be available to the user.

Limitations due to the gradient-based optimization were addressed in Black-DROPS \cite{chatzilygeroudis2017black}, which adopts a gradient-free policy optimization. In this way, also non-differentiable cost functions can be used, and the computational time can be improved with the parallelization of the black-box optimizer. With this strategy, Black-DROPS achieves similar data efficiency to PILCO's, but significantly increases asymptotic performance.

Other approaches focused on improving the accuracy of long-term predictions, overcoming approximations due to moment matching. A first attempt has been proposed in \cite{mchutchon2015nonlinear}, where long-term distributions are computed relying on particle-based methods. Based on the current policy and the one-step-ahead GP models, the authors  simulate the evolution of a batch of particles sampled from the initial state distribution. Then, the particle trajectories are used to approximate the expected cumulative cost. The policy gradient is computed using the strategy proposed in PEGASUS \cite{ng2013pegasus}, where by fixing the initial random seed, a probabilistic Markov decision process (MDP) is transformed into an equivalent partially observable MDP with deterministic transitions. Compared to PILCO, results obtained were not satisfactory. The poor performance was attributed to the policy optimization method, and in particular, to its inability to escape from the numerous local minima generated by the multimodal distribution. 

Another particle-based approach is PIPPS \cite{parmas2018pipps}, where they proposed the \emph{total propagation algorithm} to compute the gradient instead of the PEGASUS strategy. The \emph{total propagation algorithm} combines the gradient obtained with the \emph{reparameterization trick} with the \textit{likelihood ratio} gradient. The \emph{reparameterization trick} has been introduced with successful results in stochastic variational inference (SVI) \cite{kingma2013auto, rezende}. In contrast with the results obtained in SVI, where just a few samples are needed to estimate the gradient, the authors of \cite{parmas2018pipps} highlighted several issues related to the gradient computed with the \emph{reparameterization trick}, due to its exploding magnitude and random direction. \cite{parmas2018pipps} concluded that policy gradient computation via particle-based methods and the \emph{reparameterization trick} was not a feasible strategy. To overcome these issues, PIPPS relies on the \textit{likelihood ratio} gradient to regularize the gradient computed with the \emph{reparameterization trick}. The algorithm performs similarly to PILCO with some improvements in the gradient computation, and in the overall performance in the presence of additional noise.\\

\textbf{Proposed approach.} In this work, we propose an MBRL algorithm named Monte Carlo Probabilistic Inference for Learning COntrol (MC-PILCO). Like PILCO, MC-PILCO is a policy gradient algorithm, which uses GPs to describe the one-step-ahead system dynamics and relies on a particle-based method to approximate the long-term state distribution instead of using moment matching. The gradient of the expected cumulative cost w.r.t. the policy parameters is obtained by backpropagation \cite{backpropagation_hinton} on the associated stochastic computational graph, exploiting the \emph{reparameterization trick}. Differently from PIPPS, where they focused on obtaining regularized estimates of the gradient, we have interpreted the optimization problem as a stochastic gradient descent (SGD) problem \cite{SGD_Bottou10large-scalemachine}. This problem has been studied in depth in the context of neural networks, where overparameterized models are optimized using noisy estimates of the gradient \cite{DNN}. Analytical and experimental studies show that the shape of the cost function and the nonlinear activation function adopted can affect dramatically the performance of SGD algorithms \cite{DNN_min_shape,RELU1,RELU2}. Motivated by the results obtained in this field, w.r.t. the previous particle-based approaches, we considered: (i) the use of less peaked cost functions, i.e., less penalizing costs, to avoid the presence of regions where the gradient is numerically almost null. (ii) During policy optimization,  we applied dropout \cite{Jdropout_hinton} to the policy parameters, in order to improve the ability to escape from local minima and obtain better performing policies.

In addition, we propose a solution to deal with partially measurable systems which are particularly relevant in real applications, introducing MC-PILCO4PMS. Indeed, unlike simulated environments, where the state is typically assumed to be fully measurable, the state of real systems might be only partially measurable. For instance, only positions are often directly measured in real robotic systems, whereas velocities are typically computed by means of estimators, such as state observers, Kalman filters, and numerical differentiation with low-pass filters. 
In this context, during policy optimization, it is important to distinguish between the states generated by the models, which aim at describing the evolution of the real system state, and the states provided to the policy. Indeed, providing to the control policy the model predictions corresponds to assuming ability to measure directly the system state, which, as mentioned before, is not possible in the real system. 
To deal with this problem, we estimate the actual states observed in the real system by applying to the predicted states the models of both the measurement system and the online estimators, and passing these estimates to the policy during training. 
In this way, we obtain robustness w.r.t. the delays and distortions caused by online filtering. Thanks to the flexibility of our particle-based approach, it is possible to easily reproduce a wide variety of filters and state estimators, e.g., numerical differentiation, low-pass filters, Kalman filters,~etc.\\ 

\textbf{Contributions.}
We present MC-PILCO, an MBRL algorithm based on particle-based methods for long-term predictions that features cost shaping, use of dropout during policy optimization, extension to any kernel functions, and the introduction of the so called speed-integration scheme. The effectiveness of the proposed method has been ablated and shown both in simulation and on real systems. 
We considered systems with up to 12-dimensional state space that are typical dimensions for GP-based MBRL algorithms. First, the advantage of each of these features has been shown on a cart-pole swing-up benchmark and validated with statistical tests. Results show a significant increase in performance, both in terms of convergence and data efficiency, as well as the capability to handle multi-modal distributions.
Second, MC-PILCO outperforms the state-of-the-art GP-based MBRL algorithms PILCO and Black-DROPS on the same simulated cart-pole system.
Third, we validated MC-PILCO on a higher-dimensional system, by successfully learning a joint-space controller for a trajectory tracking of a simulated UR5 robotic arm. 
These results support the novel conclusion that, by properly shaping the cost function and using dropout during policy optimization, the \emph{reparameterization trick} can be used effectively in GP-based MBRL and Monte Carlo methods do not suffer of gradient estimation problems, contrary to what was asserted in the previous literature.
Furthermore, the property of using any kernel function was tested using a combination of an SE and a polynomial kernel \cite{libera2019novel}, as well as a semi-parametrical kernel \cite{romeres2019semiparametrical,romeres2016online,SP_peters}. Results obtained both in simulation and on a real Furuta pendulum show that structured kernels can increase significantly data efficiency, limiting the interaction time required to learn the tasks.


Finally, we extended the algorithm to partially measurable systems, such as most existing real systems, introducing MC-PILCO4PMS. We propose the idea of having different state estimators during model learning and policy optimization. In particular, when training the policy, it is essential to incorporate in the state predicted by the models the distortions caused by the online estimators and measurement devices in the real system. The effectiveness of this approach is validated on a simulated cart-pole and on two real systems, namely, a Furuta pendulum and a ball-and-plate system.

To recap, the main results of this paper are:
\begin{itemize}
    \item Design of MC-PILCO, a GP-based policy-gradient MBRL algorithm that relies on Monte Carlo simulation with the \emph{reparameterization trick} to update the policy;
    \item We show that by properly shaping the cost function and using dropout during policy optimization, the \emph{reparameterization trick} can be effective in policy-gradient MBRL;
    \item We analyze behaviors occurring in real setups due to filtering and state estimators, and we propose MC-PILCO4PMS, a modified version of MC-PILCO capable of dealing with partially measurable systems.\\
\end{itemize}

The article is structured as follows. In Section \ref{sec:background}, some background notions are provided: we state the general problem of model-based policy gradient methods, and present modelling approaches of dynamical systems with GPs. In Section \ref{sec:proposed_approach}, we present MC-PILCO, our proposed algorithm, detailing the policy optimization and model learning techniques adopted.  In Section \ref{sec:MC-PILCO_RS}, we discuss MC-PILCO4PMS, a variation of the proposed algorithm, specifically designed for the application to systems with partially measurable state. In Section \ref{sec:ablation}, we analyze several aspects affecting the performance of MC-PILCO, such as the cost shape, dropout, and the kernel choice. In Section \ref{sec:expeirmentSim} we validate and analyse MC-PILCO in different tests on simulated environments, while, in Section \ref{sec:experimentReal}, we refer to MC-PILCO4PMS providing a proof of concept and the results obtained on a real Furuta pendulum and a ball-and-plate system. Finally, we draw conclusions in Section \ref{sec:conclusions}.


\section{Background}\label{sec:background}
In this section, we first introduce the standard framework considered in model-based policy gradient RL methods, and then discuss how to use Gaussian Process Regression (GPR) for model learning. In the latter topic, we focus on three aspects: some background notions about GPR, the description of the model for one-step-ahead predictions, and finally, we discuss long term predictions, focusing on two possible strategies, namely, moment matching and a particle-based method.

\subsection{Model-Based policy gradient}\label{sec:problem_setting}
Consider the discrete-time system described by the unknown transition function $f(\cdot,\cdot)$,
\begin{equation*}
    \boldsymbol{x}_{t+1} = f(\boldsymbol{x}_{t}, \boldsymbol{u}_{t}) + \boldsymbol{w}_{t},
\end{equation*}
where, at each time step $t$, $\boldsymbol{x}_{t} \in \mathbb{R}^{d_{\boldsymbol{x}}}$ and $\boldsymbol{u}_{t} \in \mathbb{R}^{d_{\boldsymbol{u}}}$ are, respectively, the state and the inputs of the system, while $\boldsymbol{w}_{t} \sim \mathcal{N}(0, \Sigma_{\boldsymbol{w}})$ is an independent Gaussian random variable modeling additive noise.
The cost function $c(\boldsymbol{x}_{t})$ is defined to characterize the immediate penalty for being in state $\boldsymbol{x}_{t}$.

Inputs are chosen according to a policy function $\pi_{\boldsymbol{\theta}}: \boldsymbol{x} \mapsto \boldsymbol{u}$ that depends on the parameter vector $\boldsymbol{\theta}$.

The objective is to find the policy that minimizes the expected cumulative cost over a finite number of time steps $T$, i.e.,
\begin{equation} \label{eq:expected_cost}
    J(\boldsymbol{\theta}) = \sum_{t=0}^T \mathbb{E}_{\boldsymbol{x}_{t}}\left[c(\boldsymbol{x}_{t})\right]\text{,}
\end{equation}
with an initial state distributed according to a given $p(\boldsymbol{x}_{0})$.

A model-based approach for learning a policy consists, generally, of the succession of several trials; i.e., attempts to solve the desired task. Each trial includes three main phases:

\begin{itemize}
    \item \textit{Model Learning}: the data collected from all the previous interactions are used to build a model of the system dynamics (at the first iteration, data are collected by applying possibly random exploratory controls);
    \item \textit{Policy Update}: the policy is optimized in order to minimize the cumulative cost $J(\boldsymbol{\theta})$. The optimization algorithm iteratively approximates $J(\boldsymbol{\theta})$ by simulating the system according to the current model and policy parameters $\boldsymbol{\theta}$, and updates $\boldsymbol{\theta}$.
    \item \textit{Policy Execution}: the current optimized policy is applied to the system and the data are stored for model improvement.
\end{itemize}
Model-based policy gradient methods use the learned model to predict the state evolution when the current policy is applied. These predictions are used to estimate $J(\boldsymbol{\theta})$ and its gradient $\nabla_{\boldsymbol{\theta}}J$ in order to update the policy parameters $\boldsymbol{\theta}$ following a gradient-descent approach.

\subsection{GPR and one-step-ahead predictions}
\label{onestep_prediction}
A common strategy with GPR-based approaches consists of modeling the evolution of each state dimension with a distinct GP. Let's denote by $\Delta^{(i)}_t = x^{(i)}_{t+1} - x^{(i)}_{t}$ the difference between the value of the \emph{i}-th component at time $t+1$ and $t$, and by $y^{(i)}_{t}$ the noisy measurement of $\Delta^{(i)}_t$ with $i \in \{1,\ldots, d_{\boldsymbol{x}}\}$. Moreover, let $\tilde{\boldsymbol{x}}_{t} = [\boldsymbol{x}_{t}, \boldsymbol{u}_{t}]$ be the vector that includes the state and the input at time $t$, also called the GP input. Then, given the data $\mathcal{D}=\left(\tilde{X}, \boldsymbol{y}^{(i)}\right)$, where $\boldsymbol{y}^{(i)} = [y^{(i)}_{t_1},\dots,y^{(i)}_{t_n}]^T$ is a vector of $n$ output measurements, and $\tilde{X} = \{\tilde{\boldsymbol{x}}_{t_1},\dots,\tilde{\boldsymbol{x}}_{t_n}\}$ is the set of GP inputs, GPR assumes the following probabilistic model, for each state dimension,
\begin{equation*}
\boldsymbol{y}^{(i)} = \begin{bmatrix}
h^{(i)}(\tilde{\boldsymbol{x}}_{t_1}) \\ \vdots \\ h^{(i)}(\tilde{\boldsymbol{x}}_{t_n})
\end{bmatrix} +
\begin{bmatrix}
e^{(i)}_{t_1} \\ \vdots \\ e^{(i)}_{t_n}
\end{bmatrix}
= \boldsymbol{h}^{(i)}(\tilde{X}) + \boldsymbol{e}^{(i)} \text{,}
\end{equation*}
where $\boldsymbol{e}^{(i)}$ is a zero-mean Gaussian i.i.d. noise with standard deviation $\sigma_i$, $h^{(i)}(\cdot)$ is an unknown function modeled a priori as a zero-mean Gaussian Process, and $i \in \{1,\ldots,d_x\}$. In particular, we have $\boldsymbol{h}^{(i)} \sim \mathcal{N}(0,K_i(\tilde{X},\tilde{X}))$, with the a priori covariance matrix $K_i(\tilde{X},\tilde{X}) \in \mathbb{R}^{n\times n}$ defined element-wise through a kernel function $k_i(\cdot,\cdot)$, namely, the element in \emph{j}-th row and \emph{k}-th column is given by $k_i(\tilde{\boldsymbol{x}}_{t_j},\tilde{\boldsymbol{x}}_{t_k})$.
A crucial aspect in GPR is the kernel choice. The kernel function encodes prior assumptions about the process. One of the most common choices for continuous functions is the SE kernel, defined as
\begin{equation}\label{eq:SE}
    k_{SE}(\tilde{\boldsymbol{x}}_{t_j},\tilde{\boldsymbol{x}}_{t_k}) := \lambda^2 e^{-||\tilde{\boldsymbol{x}}_{t_j}-\tilde{\boldsymbol{x}}_{t_k}||^2_{\Lambda^{-1}}}\text{,}
\end{equation}
where the scaling factor $\lambda$ and the matrix $\Lambda$ are kernel hyperparameters which can be estimated by marginal likelihood maximization. Typically, $\Lambda$ is assumed to be diagonal, with the diagonal elements named length-scales.

Remarkably, the posterior distribution of $h^{(i)}(\cdot)$ can be computed in closed form. Let $\tilde{\boldsymbol{x}}_{t}$ be a general GP input at time $t$. Then, the distribution of $\hat{\Delta}^{(i)}_{t}$, the estimate of $\Delta^{(i)}_{t}$, is Gaussian with mean and variance given by
\begin{align}
    &\mathbb{E}[\hat{\Delta}^{(i)}_{t}] = k_i(\tilde{\boldsymbol{x}}_{t},\tilde{X})\Gamma^{-1}_i\boldsymbol{y}^{(i)} \label{eq:delta_mean} \text{,}\\
    &var[\hat{\Delta}^{(i)}_{t}] = k_i(\tilde{\boldsymbol{x}}_{t},\tilde{\boldsymbol{x}}_{t})-k_i(\tilde{\boldsymbol{x}}_{t},\tilde{X})  \Gamma^{-1}_i k_i^T(\tilde{\boldsymbol{x}}_{t},\tilde{X}) \text{,} \label{eq:delta_var}
\end{align}
with $\Gamma_i$ and $k_i(\tilde{\boldsymbol{x}}_{t},\tilde{X})$ defined as
\begin{align*}
    &\Gamma_i = (K_i(\tilde{X},\tilde{X})+\sigma_i^2I) \text{,}\\
    &k_i(\tilde{\boldsymbol{x}}_{t},\tilde{X}) = [k_i(\tilde{\boldsymbol{x}}_{t}, \tilde{\boldsymbol{x}}_{t_1}),\dots,k_i(\tilde{\boldsymbol{x}}_{t}, \tilde{\boldsymbol{x}}_{t_n})]\text{.}
\end{align*}
Recalling that the evolution of each state dimension is modeled with a distinct GP, and assuming that the GPs are conditionally independent given the current GP input $\tilde{\boldsymbol{x}}_{t}$, the posterior distribution for the estimated state at time $t+1$ is
\begin{equation}\label{eq:conditional_p}
    p(\hat{\boldsymbol{x}}_{t+1}|\tilde{\boldsymbol{x}}_{t},\mathcal{D}) \sim \mathcal{N}(\boldsymbol{\mu}_{t+1}, \Sigma_{t+1}) \text{,}
\end{equation}
where
\begin{align}
    &\boldsymbol{\mu}_{t+1} = \boldsymbol{x}_{t} + \left[\mathbb{E}[\hat{\Delta}^{(1)}_{t}],\dots, \mathbb{E}[\hat{\Delta}^{(d_{\boldsymbol{x}})}_{t}]\right]^T \text{,}\label{eq:one_step_mean}\\
    &\Sigma_{t+1} = \text{diag}\left(\left[var[\hat{\Delta}^{(1)}_{t}],\dots, var[\hat{\Delta}^{(d_{\boldsymbol{x}})}_{t}]\right]\right)\text{.}\label{eq:one_step_var}
\end{align}

\subsection{Long-term predictions with GP dynamical models}\label{subsec:longterm_prediction}
In MBRL, the policy $\pi_{\boldsymbol{\theta}}$ is evaluated and improved based on long-term predictions of the state evolution: $p(\hat{\boldsymbol{x}}_{1}), \dots, p(\hat{\boldsymbol{x}}_{T})$. The exact computation of these quantities entails the application of the one-step-ahead GP models in cascade, considering the propagation of the uncertainty. More precisely, starting from a given initial distribution $p(\boldsymbol{x}_{0})$, at each time step \emph{t}, the next state distribution is obtained by  marginalizing \eqref{eq:conditional_p} over $p(\hat{\boldsymbol{x}}_{t})$, namely,

\begin{equation}\label{long_term_integral}
    p(\hat{\boldsymbol{x}}_{t+1}) = \int  p(\hat{\boldsymbol{x}}_{t+1}|\hat{\boldsymbol{x}}_{t}, \pi_{\boldsymbol{\theta}}(\hat{\boldsymbol{x}}_{t}), \mathcal{D}) p(\hat{\boldsymbol{x}}_{t}) d\hat{\boldsymbol{x}}_{t}.
\end{equation}

Unfortunately, computing the exact predicted distribution in (\ref{long_term_integral}) is not tractable. There are different ways to solve it approximately, and here we discuss two main approaches: moment matching, adopted by PILCO, and a particle-based method, the strategy followed in this work.

\subsubsection{Moment matching}
Assuming that the GP models use only the SE kernel as a prior covariance, and considering a normal initial state distribution $x_0 \thicksim \mathcal{N}(\boldsymbol{\mu}_0,\Sigma_0)$, the first and the second moments of $p(\hat{\boldsymbol{x}}_1)$ can be computed in closed form \cite{deisenroth2013gaussian}. Then, the distribution $p(\hat{\boldsymbol{x}}_1)$ is approximated to be a Gaussian distribution, whose mean and variance correspond to the moments computed previously. Finally, the subsequent probability distributions are computed iterating the procedure for each time step of the prediction horizon. For details about the computation of the first and second moments, we refer the reader to \cite{deisenroth2013gaussian}. Moment matching offers the advantage of providing a closed-form solution for handling uncertainty propagation through the GP dynamics model. Thus, in this setting, it is possible to analytically compute the policy gradient from long-term predictions. However, as already mentioned in Section \ref{sec:introduction}, the Gaussian approximation performed in moment matching is also the cause of two main weaknesses: (i) The computation of the two moments has been performed assuming the use of SE kernels, which might lead to poor generalization properties in data that have not been seen during training \cite{ GIP, romeres2019semiparametrical,romeres2016online,SP_peters}. (ii) Moment matching allows modeling only unimodal distributions, which might be a too restrictive approximation of the real system behavior.

\subsubsection{Particle-based method}\label{particlesMethod}
\begin{figure}
\centering
  \includegraphics[width=0.75\linewidth]{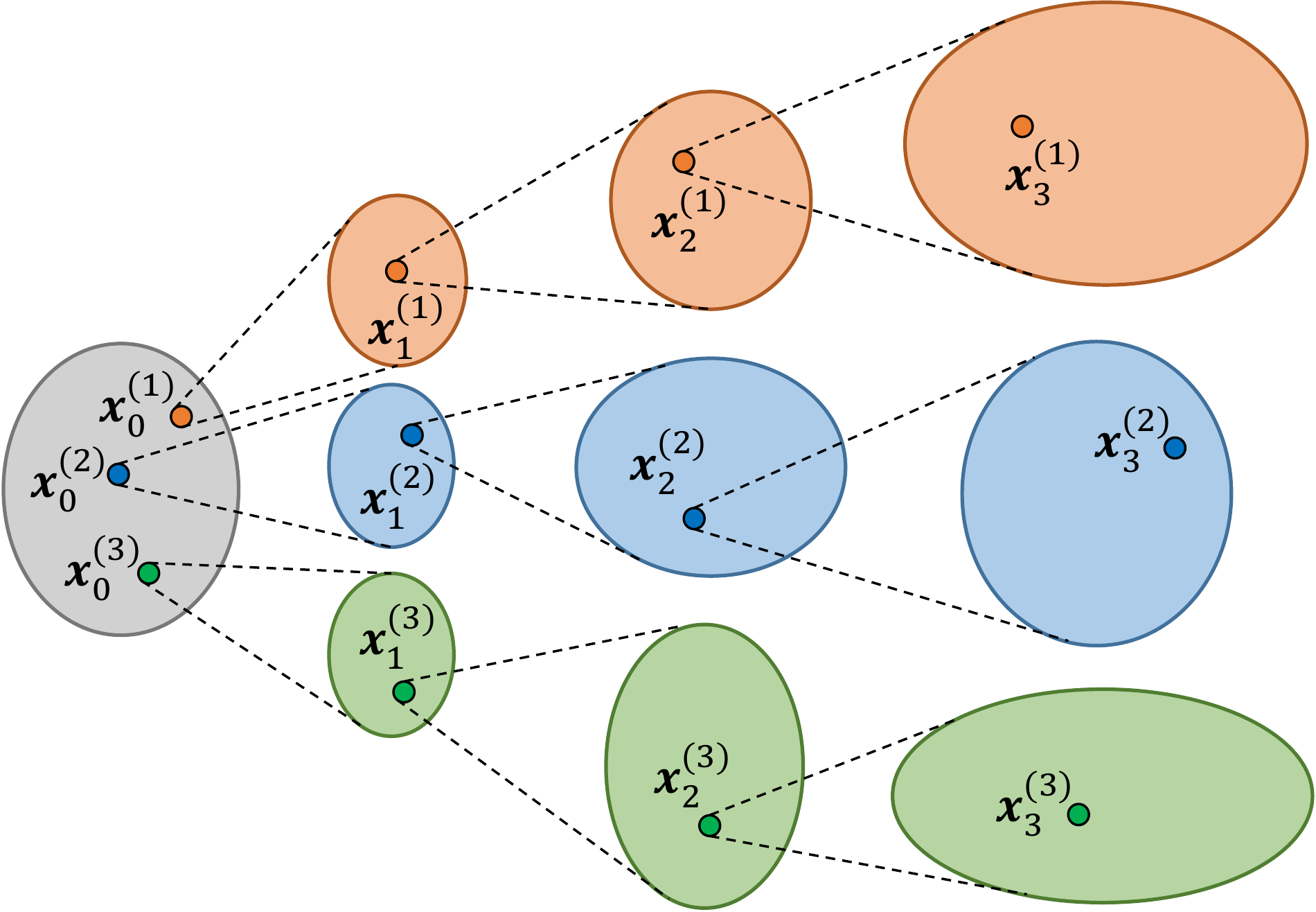}
  \caption{\small Example of three particles propagating through the stochastic model (Gaussian distributions represented as ellipses).}
  \label{fig:particles}
\end{figure}

The integral in \eqref{long_term_integral} can be approximated relying on Monte Carlo approaches, in particular on particle-based methods, see, for instance, \cite{chatzilygeroudis2017black} \cite{parmas2018pipps}. Specifically, $M$ particles are sampled from the initial state distribution $p(\boldsymbol{x}_0)$. Each one of the $M$ particles is propagated using the one-step-ahead GP models \eqref{eq:conditional_p}. Let $\boldsymbol{x}^{(m)}_t$ be the state of the \emph{m}-th particle at time $t$, with $m=1,\dots,M$. At time step $t$, the actual policy $\pi_{\boldsymbol{\theta}}$ is evaluated to compute the associated control. The GP model provides the Gaussian distribution $p\left(\boldsymbol{x}^{(m)}_{t+1}|\boldsymbol{x}^{(m)}_t,\pi_{\boldsymbol{\theta}}(\boldsymbol{x}^{(m)}_t),\mathcal{D}\right)$ from which $\boldsymbol{x}^{(m)}_{t+1}$, the state of the particle at the next time step, is sampled. This process is iterated until a trajectory of length $T$ is generated for each particle. The overall process is illustrated in Figure \ref{fig:particles}. The long-term distribution at each time step is approximated with the distribution of the particles. Note that this approach does not impose any constraint on the choice of the kernel function and the initial state distribution. Moreover, there are no restrictions on the distribution of $p(\hat{\boldsymbol{x}}_t)$. Therefore, particle-based methods do not suffer from the problems seen in moment matching, at the cost of being more computationally heavy. Specifically, the computation of \eqref{eq:conditional_p} entails the computation of \eqref{eq:delta_mean} and \eqref{eq:delta_var}, which are, respectively, the mean and the variance of the delta states. Regarding the computational complexity, it can be noted that $\Gamma^{-1}_i\boldsymbol{y}^{(i)}$ is computed a single time offline during the training of the GP model (same computation is needed in the moment matching case), and the number of operations required to compute \eqref{eq:delta_mean} is linear w.r.t. the number of samples $n$. The computational bottleneck is the computation of \eqref{eq:delta_var}, which is $O(n^2)$. Then, the cost of a single state prediction is $O(d_{\boldsymbol{x}} n^2)$, leading to a total computational cost of $O(d_{\boldsymbol{x}} MTn^2)$. Depending on the complexity of the system dynamics, the number of particles necessary to obtain a good approximation might be high, determining a considerable computational burden. Nevertheless, the computational burden can be substantially mitigated via GPU parallel computing, due to the possibility of computing the evolution of each particle in parallel.

\section{MC-PILCO}\label{sec:proposed_approach}
In this section, we present the proposed algorithm. MC-PILCO relies on GPR for model learning and follows a Monte Carlo sampling method to estimate the expected cumulative cost from particles trajectories propagated through the learned model. We exploit the \emph{reparameterization trick} to obtain the policy gradient from the sampled particles and optimize the policy. This way of proceeding is very flexible, and allows using any kind of kernels for the GPs, as well as providing more reliable approximations of the system's behaviour.
MC-PILCO, in broad terms, consists of the iteration of three main steps, namely, update the GP models, update the policy parameters, and execute the policy on the system. In its turn, the policy update is composed of the following three steps, iterated for a maximum of $N_{opt}$ times:
\begin{itemize}
    \item simulate the evolution of $M$ particles, based on the current $\pi_{\boldsymbol{\theta}}$ and on the GP models learned from the previously observed data;
    \item compute $\hat{J}(\boldsymbol{\theta})$, an approximation of the expected cumulative cost, based on the evolution of the $M$ particles;
    \item update the policy parameters $\boldsymbol{\theta}$ based on $\nabla_{\boldsymbol{\theta}}\hat{J}(\boldsymbol{\theta})$, the gradient of $\hat{J}(\boldsymbol{\theta})$ w.r.t. $\boldsymbol{\theta}$, computed by backpropagation.
\end{itemize}
In the remainder of this section, we discuss in greater depth the model learning step and the policy optimization step.

\subsection{Model Learning}\label{sec:modelLearning}
Here, we describe the model learning framework considered in MC-PILCO. We begin by showing the proposed one-step-ahead prediction model, and analyzing the advantages w.r.t. the standard model described in Section \ref{onestep_prediction}. Then, we discuss the choice of the kernel functions. Finally, we briefly detail the model's hyperparameters optimization and the strategy adopted to reduce the computational cost.

\subsubsection{Speed-integration model}
Let the state be defined as $\boldsymbol{x}_{t} = [\boldsymbol{q}_{t}^T,\boldsymbol{\dot{q}}_{t}^T]^T$, where $\boldsymbol{q}_{t}\in \mathbb{R}^{\nicefrac{d_{\boldsymbol{x}}}{2}}$ is the vector of the generalized coordinates of the system at time step $t$, and, $\boldsymbol{\dot{q}}_{t}$ represents the derivative of $\boldsymbol{q}_{t}$ w.r.t. time. MC-PILCO adopts a one-step-ahead  model, hereafter denoted as \textit{speed-integration} dynamical model, which exploits the intrinsic correlation between the state components $\boldsymbol{q}$ and $\boldsymbol{\dot{q}}$. Indeed, when considering a sufficiently small sampling time $T_s$ (small w.r.t. the application), it is reasonable to assume constant accelerations between two consecutive time-steps, obtaining the following evolution of~$\boldsymbol{q}_t$,
\begin{equation}\label{eq:delta_pos_const}
    \boldsymbol{q}_{t+1} = \boldsymbol{q}_t + T_s \boldsymbol{\dot{q}}_t + \frac{T_s}{2}(\boldsymbol{\dot{q}}_{t+1}- \boldsymbol{\dot{q}}_t)\text{.}
\end{equation}
Let $\mathcal{I}_{\boldsymbol{{q}}}$ (respectively $\mathcal{I}_{\boldsymbol{\dot{q}}}$) be the ordered set of the dimension indices of the state $\boldsymbol{x}$ associated with $\boldsymbol{{q}}$ (respectively $\boldsymbol{\dot{q}}$ ). The proposed \textit{speed-integration} model learns only $d_{\boldsymbol{x}}/2$ GPs, each of which models the evolution of a distinct velocity component $\Delta^{(i_k)}_{t}$, with $i_k\in \mathcal{I}_{\boldsymbol{\dot{q}}}$. Then, the evolution of the position change, $\Delta^{(i_k)}_{t}$, with $i_k\in \mathcal{I}_{\boldsymbol{q}}$, is computed according to \eqref{eq:delta_pos_const} and the predicted change in velocity.

Many previous MBRL algorithms, see for instance  \cite{deisenroth2011pilco,chatzilygeroudis2017black}, adopted the standard model described in Section \ref{onestep_prediction}, and hereafter denoted as \textit{full-state} dynamical model. The \textit{full-state} model predicts the change of each state component with a distinct and independent GP. Doing so, the evolution of each state dimension is assumed to be conditionally independent given the current GP input, and it is necessary to learn a number of GPs equal to the state dimension $d_{\boldsymbol{x}}$. Then, compared to the \textit{full-state} model, the  proposed \textit{speed-integration} model halves the number of GPs to be learned, decreasing the cost of a state prediction to $O( \frac{d_{\boldsymbol{x}}}{2} MTn^2)$. Nevertheless, this approach is based on a constant acceleration assumption, and works properly only when considering small enough sampling times. However, MC-PILCO can use also the standard \textit{full-state} model, which might be more effective when sampling time is longer.

\subsubsection{Kernel functions}\label{sec:kernels}
Regardless of the GP dynamical model structure adopted, one of the advantages of the particle-based policy optimization method is the possibility of choosing any kernel functions without restrictions. Hence, we considered different kernel functions as examples to model the evolution of physical systems. However, readers can consider a custom kernel function appropriate for their application.

\textbf{Squared exponential (SE)}. The SE kernel described in \eqref{eq:SE} represents the standard choice adopted in many different works.

\textbf{SE + Polynomial (SE+$\text{P}^{(d)}$)}. Recalling that the sum of kernels is still a kernel \cite{williams2006gaussian}, we considered also a function given by the sum of a SE and a polynomial kernel. In particular, we used the Multiplicative Polynomial (MP) kernel, which is a refinement of the standard polynomial kernel, introduced in \cite{libera2019novel}. The MP kernel of degree $d$ is defined as the product of $d$ linear kernels, namely,
\begin{equation*}
    k_{P}^{(d)}(\tilde{\boldsymbol{x}}_{t_j},\tilde{\boldsymbol{x}}_{t_k}) := \prod_{r=1}^ d\left(\sigma^2_{P_r} + \tilde{\boldsymbol{x}}_{t_j}^T\Sigma_{P_r} \tilde{\boldsymbol{x}}_{t_k}\right)\text{.}
\end{equation*}
where the $\Sigma_{P_r}>0$ matrices are distinct diagonal matrices. The diagonal elements of the $\Sigma_{P_r}$, together with the $\sigma^2_{P_r}$ elements are the kernel hyperparameters. The resulting kernel is
\begin{equation}\label{eq:SE+Pkernel}
    k_{SE+P^{(d)}}(\tilde{\boldsymbol{x}}_{t_j},\tilde{\boldsymbol{x}}_{t_k}) =k_{SE}(\tilde{\boldsymbol{x}}_{t_j},\tilde{\boldsymbol{x}}_{t_k}) + k_{P}^{(d)}(\tilde{\boldsymbol{x}}_{t_j},\tilde{\boldsymbol{x}}_{t_k})\text{.}
\end{equation}
The idea motivating this choice is the following: the MP kernel allows capturing possible modes of the system that are polynomial functions in $\tilde{\boldsymbol{x}}$, which are typical in mechanical systems \cite{GIP}, while the SE kernel models more complex behaviors not captured by the polynomial kernel.

\textbf{Semi-Parametrical (SP)}. When prior knowledge about the system dynamics is available, for example given by physics first principles, the so called physically inspired (PI) kernel can be derived. The PI kernel is a linear kernel defined on suitable basis functions $\phi(\tilde{\boldsymbol{x}})$, see for instance \cite{romeres2019semiparametrical}. More precisely, $\boldsymbol{\phi}(\tilde{\boldsymbol{x}}) \in \mathbb{R}^{d_{\phi}}$ is a (possibly nonlinear) transformation of the GP input $\tilde{\boldsymbol{x}}$ determined by the physical model. Then, we have
\begin{equation*}
    k_{PI}(\tilde{\boldsymbol{x}}_{t_j},\tilde{\boldsymbol{x}}_{t_k}) = \boldsymbol{\phi}^T(\tilde{\boldsymbol{x}}_{t_j}) \Sigma_{PI}\boldsymbol{\phi}(\tilde{\boldsymbol{x}}_{t_k}) \text{,}
\end{equation*}
where $\Sigma_{PI}$ is a $d_{\phi} \times d_{\phi}$ positive-definite matrix, whose elements are the $k_{PI}$ hyperparameters; to limit the number of hyperparameters, a standard choice consists in considering $\Sigma_{PI}$ to be diagonal. To compensate possible inaccuracies of the physical model, it is common to combine $k_{PI}$ with an SE kernel, obtaining so called semi-parametric kernels \cite{SP_peters, romeres2019semiparametrical}, expressed as
\begin{equation*}\label{eq:SP_kernel}
    k_{SP}(\tilde{\boldsymbol{x}}_{t_j},\tilde{\boldsymbol{x}}_{t_k}) = k_{PI}(\tilde{\boldsymbol{x}}_{t_j},\tilde{\boldsymbol{x}}_{t_k}) + k_{SE}(\tilde{\boldsymbol{x}}_{t_j},\tilde{\boldsymbol{x}}_{t_k}) \text{.}
\end{equation*}
The rationale behind this kernel is the following: $k_{PI}$ encodes the prior information given by the physics, and $k_{SE}$ compensates for the dynamical components unmodeled in $k_{PI}$.

\subsubsection{Model optimization and reduction techniques}\label{subsec:GPoptimization} In MC-PILCO, the GP hyperparameters are optimized by maximizing the  marginal likelihood (ML) of the training samples, see \cite{williams2006gaussian}. In Section \ref{particlesMethod}, we saw that the computational cost of a particle prediction scales with the square of the number of samples $n$, leading to a considerable computational burden when $n$ is high. In this context, it is essential to implement a strategy to limit the computational complexity of a prediction. 
We implemented a \textit{Subset of Data} technique (refer to \cite{GP_approx_overview} for further details on this method and others) with an input selection procedure inspired by \cite{sparse_online_GP}, where the authors proposed an online importance sampling strategy. After optimizing the GP hyperparameters by ML maximization, the samples in $\mathcal{D}$ are downsampled to a subset $\mathcal{D}_r = \left(\tilde{X}_r,\boldsymbol{y}^{(i)}_r\right)$, which is then used to compute the predictions. This procedure first initializes $\mathcal{D}_r$ with the first sample in $\mathcal{D}$, then, it computes iteratively the GP estimates of all the remaining samples in $\mathcal{D}$, using $\mathcal{D}_r$ as training samples. Each sample in $\mathcal{D}$ is either added to $\mathcal{D}_r$ if the uncertainty of the estimate is higher than a threshold $\beta^{(i)}$ or it is discarded. The GP estimator is updated every time a sample is added to $\mathcal{D}_r$. The trade-off between the reduction of the computational burden and the severity of the approximation introduced is regulated by tuning $\beta^{(i)}$. The higher the $\beta^{(i)}$, the smaller the number of samples in $\mathcal{D}_r$. Inversely, using values of $\beta^{(i)}$ that are too high might compromise the accuracy of GP predictions. 

\subsection{Policy optimization}\label{subsec:MCPILCO_policy_opt}
\begin{figure*}[thpb]
	\centering
	\begin{subfigure}[b]{0.49\textwidth}
         \centering
         \includegraphics[width=\textwidth]{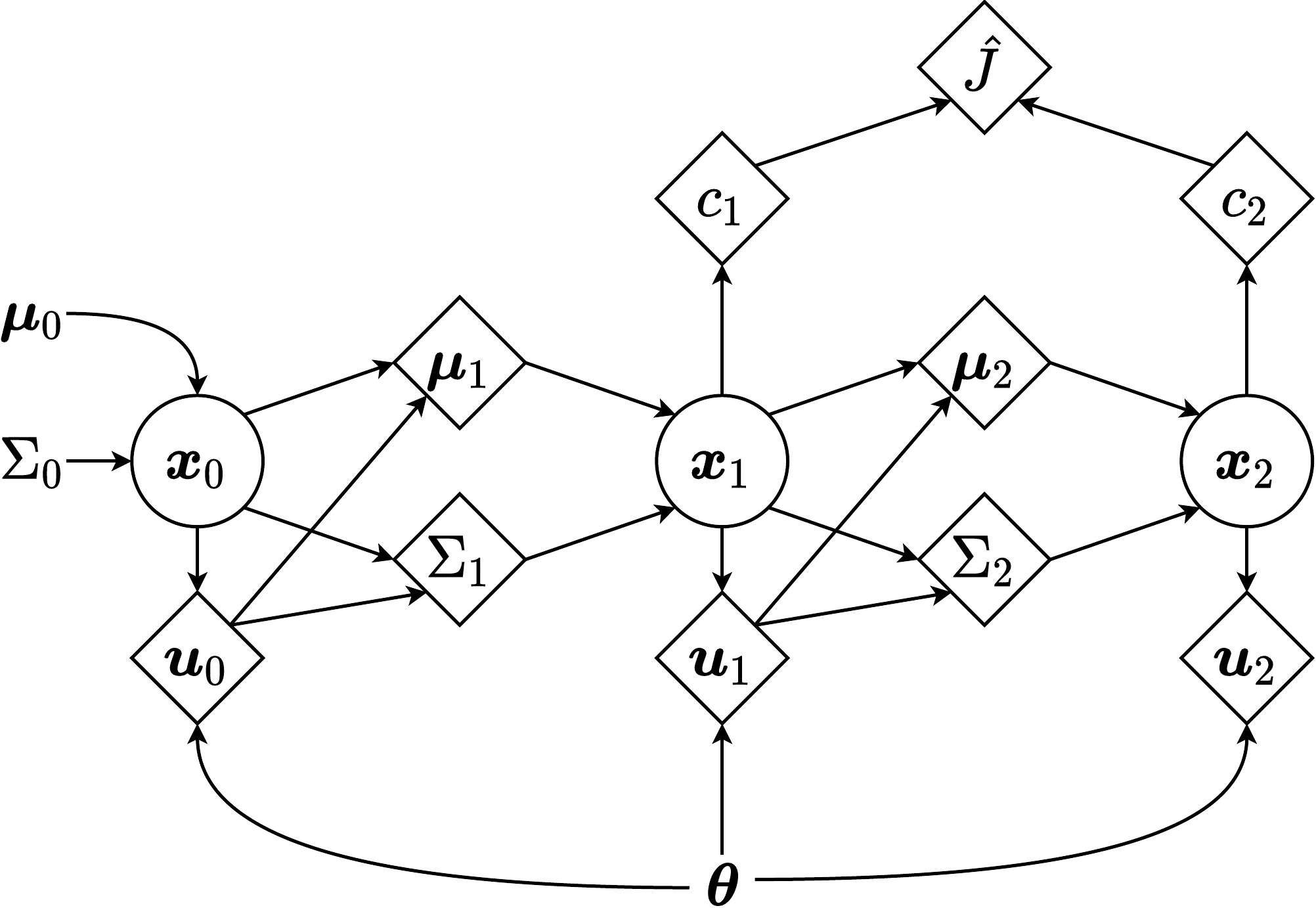}
         \caption{Original computational graph.}
     \end{subfigure}
     \hfill
	\begin{subfigure}[b]{0.49\textwidth}
         \centering
         \includegraphics[width=\textwidth]{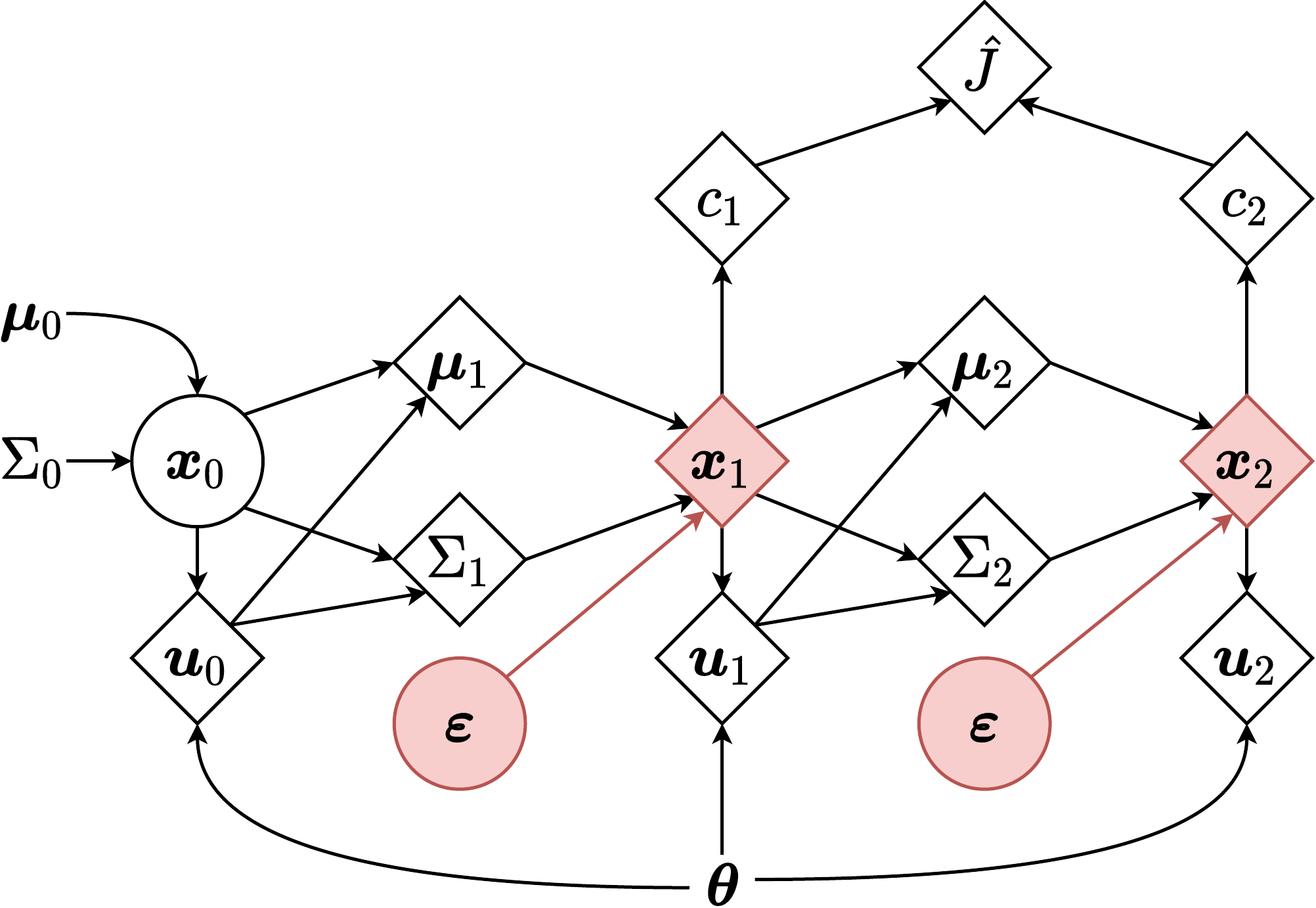}
         \caption{Reparameterized computational graph.}
     \end{subfigure}
	\caption{(Left) Original computational graph of the GP model predictions for two time steps. (Right) Computational graph modified by the \emph{reparameterization trick}. Squares and circles represent, respectively, deterministic and stochastic operations.}
	\label{fig:rep_trick}
\end{figure*}
Here, we present the policy optimization strategy adopted in MC-PILCO. We start by describing the general-purpose policy structure considered. Later, we show how to exploit backpropagation and the \emph{reparameterization trick} to estimate the policy gradient from particle-based long-term predictions. Finally, we explain how to implement dropout in this framework.

\subsubsection{Policy structure}
In all the experiments presented in this work, we adopted an RBF network policy with outputs limited by an hyperbolic tangent function, properly scaled. We call this function \textit{squashed-RBF-network}, and it is defined as
\begin{equation}\label{eq:policy}
    \pi_{\boldsymbol{\theta}}(\boldsymbol{x}) = u_{max}\;\text{tanh} \left(\frac{1}{u_{max}}\sum_{i=1}^{n_b} w_i e^{||\boldsymbol{a}_i-\boldsymbol{x}||_{\Sigma_{\pi}}^2}\right)\text{.}
\end{equation}
The policy parameters are $\boldsymbol{\theta} = \left\{\boldsymbol{w}, A,\Sigma_{\pi}\right\}$, where $\boldsymbol{w}=[w_1\dots w_{n_b}]$ and $A=\left\{\boldsymbol{a}_1\dots\boldsymbol{a}_{n_b}\right\}$ are, respectively, the weights and the centers of the Gaussian basis functions, while ${\Sigma_{\pi}}$ determines the shape of the Gaussian basis functions; in all experiments we assumed ${\Sigma_{\pi}}$ to be diagonal. The maximum control action $u_{max}$ is constant and chosen depending on the system to control. It is worth mentioning that MC-PILCO can deal with any differentiable policy, so more complex functions, such as deep neural networks, could be considered too.

\subsubsection{Policy gradient estimation}
MC-PILCO derives the policy gradient by applying the \emph{reparameterization trick} to the computation of the estimated expected cumulative cost in \eqref{eq:expected_cost}, obtained relying on Monte Carlo sampling \cite{caflisch1998monte}.
Given a control policy $\pi_{\boldsymbol{\theta}}$ and an initial state distribution $p(\boldsymbol{x}_0)$, the evolution of a sufficiently high number of particles is predicted as described in Section \ref{particlesMethod}. Thus, the sample mean of the costs incurred by the particles at time step $t$ approximates each $\mathbb{E}_{\boldsymbol{x}_t}[c(\boldsymbol{x}_t)]$.
Specifically, let $\boldsymbol{x}^{(m)}_t$ be the state of the \emph{m}-th particle at time $t$, with $m=1,\dots,M$ and $t=0,\dots,T$. The Monte Carlo estimate of the expected cumulative cost is computed with the following expression:
\begin{equation}\label{eq:est_J}
    \hat{J}(\boldsymbol{\theta}) = \sum_{t=0}^T \left( \frac{1}{M}\sum_{m=1}^M c\left(\boldsymbol{x}_t^{(m)}\right)\right) \;.
\end{equation}
The evolution of every particle $\boldsymbol{x}^{(m)}_t$ at the next time step is sampled from the normal distribution  $p(\boldsymbol{x}^{(m)}_{t+1}|\boldsymbol{x}^{(m)}_t,\pi_{\boldsymbol{\theta}}(\boldsymbol{x}^{(m)}_t),\mathcal{D}) \sim \mathcal{N}(\boldsymbol{\mu}_{t+1}, \Sigma_{t+1})$, defined in (\ref{eq:one_step_mean})-(\ref{eq:one_step_var}). Hence, the computation of $\hat{J}(\boldsymbol{\theta})$ entails the sampling from probability distributions that depend on policy parameters $\boldsymbol{\theta}$. The presence of such stochastic operations makes it impossible to compute straightforwardly the gradient of \eqref{eq:est_J} w.r.t. the policy parameters.
The \emph{reparameterization trick} \cite{kingma2013auto} allows to still differentiate through the stochastic operations by re-defining the probability distributions involved in the computation of $\nabla_{\boldsymbol{\theta}}\hat{J}$. In fact, instead of sampling directly from $\mathcal{N}(\boldsymbol{\mu}_{t+1}, \Sigma_{t+1})$, it is possible to sample a point $\epsilon$ from a zero-mean and unit-variance normal distribution with the same dimension of $\boldsymbol{\mu}_{t+1}$. Then, $\epsilon$ can be mapped into the desired distribution as $\boldsymbol{x}^{(m)}_{t+1}=\boldsymbol{\mu}_{t+1}+L_{t+1}\epsilon$, where $L_{t+1}$ is the Cholesky decomposition of $\Sigma_{t+1}$, namely, $\Sigma_{t+1}=L_{t+1}L^T_{t+1}$. In this way, the \emph{reparameterization trick} makes the dependency of $\boldsymbol{x}^{(m)}_{t+1}$ from $\boldsymbol{\theta}$ purely deterministic, allowing to compute $\nabla_{\boldsymbol{\theta}}\hat{J}$ simply by backpropagation. Figure \ref{fig:rep_trick} illustrates how the \emph{reparameterization trick} works in the context of MC-PILCO. Then, policy parameters $\boldsymbol{\theta}$ are updated using the Adam solver \cite{kingma2014adam}; we will denote the Adam step size with $\alpha_{lr}$.

\subsubsection{Dropout}
To improve exploration in the parameter space and increase the ability of escaping from local minima during policy optimization, we considered the use of dropout \cite{Jdropout_hinton}. The adopted procedure is described assuming that the policy is the \textit{squashed-RBF-network} in \eqref{eq:policy}; similar considerations can be applied to different policy functions. When dropout is applied to the policy in \eqref{eq:policy}, weights $\boldsymbol{w}$ are randomly dropped with probability $p_d$ at each evaluation of the policy. This operation is performed by scaling each weight $w_i$ with a random variable $r_i\sim Bernoulli(1-p_d)$, where $Bernoulli(1-p_d)$ denotes a Bernoulli distribution, assuming value $1/(1-p_d)$ with probability $1-p_d$, and $0$ with probability $p_d$. This operation is equivalent to defining a probability distribution for $\boldsymbol{w}$, obtaining a parameterized stochastic policy. In particular, as shown in \cite{dropout_ghahramani}, the distribution of each $w_i$ can be approximated with a bimodal distribution, defined by the sum of two properly scaled Gaussian distributions with infinitely small variance $\xi^2$, namely,
\begin{equation*}
    p_d \mathcal{N}(0,\xi^2) + (1-p_d) \mathcal{N}\left(\frac{w_i}{1-p_d}, \xi^2\right)\text{.}
\end{equation*}
The use of a stochastic policy during policy optimization allows increasing the entropy of the particles' distribution. This property increments the probability of visiting low-cost regions and escaping from local minima. In addition, we also verified that dropout can mitigate issues related to exploding gradients. This is probably due to the fact that the average of several different values of $\boldsymbol{w}$ is used to compute the gradient and not a single value of $\boldsymbol{w}$, i.e., different policy functions are used, obtaining a regularization of the gradient estimates.

By contrast, the use of a stochastic policy might affect the precision of the obtained solution due to the additional entropy. We also need to take into consideration that the final objective is to obtain a deterministic policy. For these reasons, we designed an heuristic scaling procedure to gradually decrease the dropout rate, $p_d$, until it equals $0$. The scaling action is triggered by a monitoring signal $s$, defined from the statistics of the past history of $\hat{J}$. Define the cost change, $\Delta\hat{J}_j = \hat{J}(\boldsymbol{\theta}_j)-\hat{J}(\boldsymbol{\theta}_{j-1})$, where $\boldsymbol{\theta}_j$ denotes the policy parameters at the \emph{j}-th optimization step. Then, $s$ is computed as a filtered version of the ratio between $\mathcal{E}[\Delta\hat{J}_j]$ and $\sqrt{\mathcal{V}[\Delta\hat{J}_j]}$, that are, respectively, the mean and the standard deviation of $\Delta\hat{J}_j$ computed with an Exponential Moving Average (EMA) filter. The expression of $s$ at the \emph{j}-th optimization step is the following:
\begin{align}
&\mathcal{E}[\Delta\hat{J}_j] = \alpha_s \mathcal{E}[\Delta\hat{J}_{j-1}] + (1-\alpha_s)\Delta\hat{J}_{j} \text{,} \nonumber\\
&\mathcal{V}[\Delta\hat{J}_j] = \alpha_s (\mathcal{V}[\Delta\hat{J}_{j-1}] + (1-\alpha_s)(\Delta\hat{J}_{j}-\mathcal{E}[\Delta\hat{J}_{j-1}])^2)\text{,}\nonumber\\
&s_j = \alpha_s s_{j-1} + (1-\alpha_s)\frac{\mathcal{E}[\Delta\hat{J}_j]}{\sqrt{\mathcal{V}[\Delta\hat{J}_j]}}\text{,} \label{eq:update_s}
\end{align}
with $\alpha_s$ a coefficient of the exponential moving average filter, which determines the memory of the filter. At each iteration of the optimization procedure, the algorithm checks if the absolute value of the monitoring signal $s$  in the last $n_s$ iterations is below the threshold $\sigma_s$, namely,
\begin{equation}\label{eq:check_s}
    [|s_{j-n_s}|\dots |s_j|] < \sigma_{s}\text{,}
\end{equation}
where $<$ is an element-wise operator, and the condition in \eqref{eq:check_s} is true if it is verified for all the elements. If the condition is verified, $p_d$ is decreased by the quantity $\Delta p_d$, and both the learning rate of the optimizer, $\alpha_{lr}$, and $\sigma_s$, are scaled by an arbitrary factor $\lambda_s$. Then, we have
\begin{subequations}\label{eq:update_opt_par}
\begin{align}
&p_d = p_d-\Delta p_d\label{eq:update_p_d}\text{,}\\
&\alpha_{lr}= \lambda_{s}\alpha_{lr}\text{,}\label{eq:update_lr}\\
&\sigma_s = \lambda_{s}\sigma_s\text{.}\label{eq:update_sigma_s}
\end{align}
\end{subequations}

The procedure is iterated as long as
\begin{equation}\label{eq:exit_condition}
    p_d\geq0 \text{ and } \alpha_{lr}\geq\alpha_{lr_{min}}\text{,}
\end{equation}
where $\alpha_{lr_{min}}$ is the minimum value of the learning rate.

The rationale behind this heuristic scaling procedure is the following. The $s_j$ signal is small, if $\mathcal{E}[\Delta\hat{J}_j]$ is close to zero, or if $\mathcal{V}[\Delta\hat{J}_j]$ is particularly high. The first case happens when the optimization reaches a minimum, while the high variance denotes that the particles' trajectories cross regions of the workspace where the uncertainty of the GPs predictions is high. In both cases, we are interested in testing the policy on the real system, in the first case to verify if the configuration reached solves the task, and in the second case to collect data where predictions are uncertain, and so to improve model accuracy. MC-PILCO is summarized in pseudo-code in Algorithm \ref{alg:MC-PILCo}.

We conclude the discussion about policy optimization by reporting, in Table \ref{tab: standard_opt_setup}, the optimization parameters used in all the proposed experiments, unless expressly stated otherwise. However, it is worth mentioning that some adaptation could be needed in other setups, depending on the problem considered.


\begin{table}[ht]
\centering
\begin{tabular}{|l|l|l|}
\hline
Parameter & Description & Value \\ \hline
$p_d$ & \textit{dropout probability} & 0.25 \\ \hline
$\Delta p_d$ & $p_d$ \textit{reduction coeff.} & 0.125  \\ \hline
$\alpha_{lr}$ & \textit{Adam step size} & 0.01  \\ \hline
$\alpha_{lr_{min}}$ & \textit{minimum step size} & 0.0025  \\ \hline
$\alpha_s$ & \textit{EMA filter coeff.} & 0.99  \\ \hline
$\sigma_{s}$ & \textit{monitoring signal treshold} & 0.08  \\ \hline
$n_s$ & \textit{num. iterations monitoring} & 200  \\ \hline
$\lambda_s$ & $\sigma_{s}$ \textit{reduction coeff.} & 0.5  \\ \hline
$M$ & number of particles & 400  \\ \hline
\end{tabular}
\caption{\small Standard values for the policy optimization parameters.}
\label{tab: standard_opt_setup}
\end{table}

\begin{algorithm}
\SetAlgoLined
 \textbf{init} policy $\pi_{\boldsymbol{\theta}}(\cdot)$, cost $c(\cdot)$, kernel $k(\cdot,\cdot)$, maximum optimization steps $N_{opt}$, number of particles $M$, learning rate $\alpha_{lr}$, min. learning rate $\alpha_{lr_{min}}$, dropout probability $p_d$, dropout probability reduction $\Delta_{p_d}$ and other monitoring signal parameters: $\sigma_s$, $\lambda_s$, $n_s$.\\
 Apply exploratory control to system and collect data\\
 \While{task not learned}{
 \textbf{1) Model Learning:}\\
  Learn GP models from sampled data - Sec.~\ref{sec:modelLearning}\;
  \textbf{2) Policy Update:}\\
  Initialize monitoring signal $s_0=0$\;
  \For{$j=1...N_{opt}$}{
  Simulate $M$ particles rollouts with GP models and current policy $\pi_{\boldsymbol{\theta}_j}(\cdot)$\;
  Compute $\hat{J}(\boldsymbol{\theta}_j)$ from particles (\ref{eq:est_J})\;
  Compute $\nabla_{\boldsymbol{\theta}}\hat{J}(\boldsymbol{\theta}_j)$ through backpropagation\;
  Gradient-based policy update $\rightarrow \pi_{\boldsymbol{\theta}_{j+1}}(\cdot)$ \;
  Update monitoring signal $s_j$ with \eqref{eq:update_s}\;
  \If{\eqref{eq:check_s} is True}{
    Update $p_d$, $\alpha_{lr}$ and $\sigma_s$ with \eqref{eq:update_opt_par};
  }
  \If{\eqref{eq:exit_condition} is False}{
    \textbf{break}\;
  }
  }
  \textbf{3) Policy Execution:}\\
  apply updated policy to system and collect data
 }
 \textbf{return} trained policy, learned GP model\;
 \caption{MC-PILCO}
 \label{alg:MC-PILCo}
\end{algorithm}

\section{MC-PILCO for Partially Measurable Systems}\label{sec:MC-PILCO_RS}
In this section, we discuss the application of MC-PILCO to systems where the state is partially measurable, i.e., systems whose state is observable, but only some components of the state can be directly measured, while the rest must be estimated from measurements. For simplicity, we introduce the problem discussing the case of a mechanical system where only positions (and not velocities) can be measured, but similar considerations can be done for any partially measurable system with observable state. Then, we describe \emph{MC-PILCO for Partially Measurable Systems} (MC-PILCO4PMS), a modified version of MC-PILCO, proposed to deal with such setups. 

Consider a mechanical systems where only joint positions can be measured. This can be described as a partially measurable system, where in the state $\boldsymbol{x}_t = [\boldsymbol{q}_t^T,\boldsymbol{\dot{q}}_t^T]^T$ only $\boldsymbol{q}_t$ is measured. Consequently, the $\boldsymbol{\dot{q}}_t$ elements are estimated starting from the history of $\boldsymbol{q}_t$ measurements through proper estimation procedures, possibly performing also denoising operations of $\boldsymbol{q}_t$ in case that the measurement noise is high. In particular, it is worth distinguishing between estimates computed online and estimates computed offline. The former are provided to the control policy to determine the system control input, and they need to respect real-time constraints, namely, velocity estimates are causal and computations must be performed within a given interval. For the latter, we do not have to deal with such constraints. As a consequence, offline estimates can be more accurate, taking into account acausal information and limiting delays and distortions.

In this context, we verified that, during policy optimization, it is relevant to distinguish between the particle state predictions computed by the models and the data provided to the policy. On the one hand, GPs should simulate the real system dynamics, independently of additional noise given by the sensing instrumentation, they need to work with the most accurate estimates available, possibly obtained with acausal filters; delays and distortions might compromise the accuracy of long-term predictions. On the other hand, providing to the policy directly the particle states computed with the GPs during policy optimization, correspond to train the policy assuming to access directly to the system state, which is not possible in the considered setup. Indeed, relevant discrepancies between the particle states and the state estimates computed online, during the interaction with the real system, might compromise the effectiveness of the policy. Most of the previous GP-based MBRL algorithms do not focus on these aspects, and assume direct access to the state. In our opinion, a correct understanding of the state estimation problem, for both modeling and control purposes, is fundamental for a robust deployment of MBRL solutions to real-world applications.

To deal with the above issues, we introduce MC-PILCO4PMS an extension of MC-PILCO, that carefully takes into account the presence of online state estimators during policy training. With respect to the algorithm described in Section \ref{sec:proposed_approach}, we propose the two following additions:\\

\textbf{Offline estimation of GPs training data.} We compute the state estimates used to train the GP models with offline estimation techniques. In particular, in our real experiments, we considered two options,
\begin{itemize}
    \item Computation of the velocities with the central difference formula, i.e., $\dot{\boldsymbol{q}}_t = (\boldsymbol{q}_{t+1}-\boldsymbol{q}_{t-1})/(2 T_s)$, where $T_s$ is the sampling time. This technique can be used only when the measurement noise is limited, otherwise the $\dot{\boldsymbol{q}}$ estimates might be too noisy.
    \item Estimation of the state with a Kalman smoother \cite{einicke2006optimal}, with state-space model given by the general equations relating positions, velocities, and accelerations. The advantage of this technique is that it exploits the correlation between positions and velocities, increasing regularization.
\end{itemize}
\vspace{3mm}

\begin{figure}[t]
     \centering
     \begin{subfigure}[t]{0.95\linewidth}
         \centering
         \includegraphics[width=\textwidth]{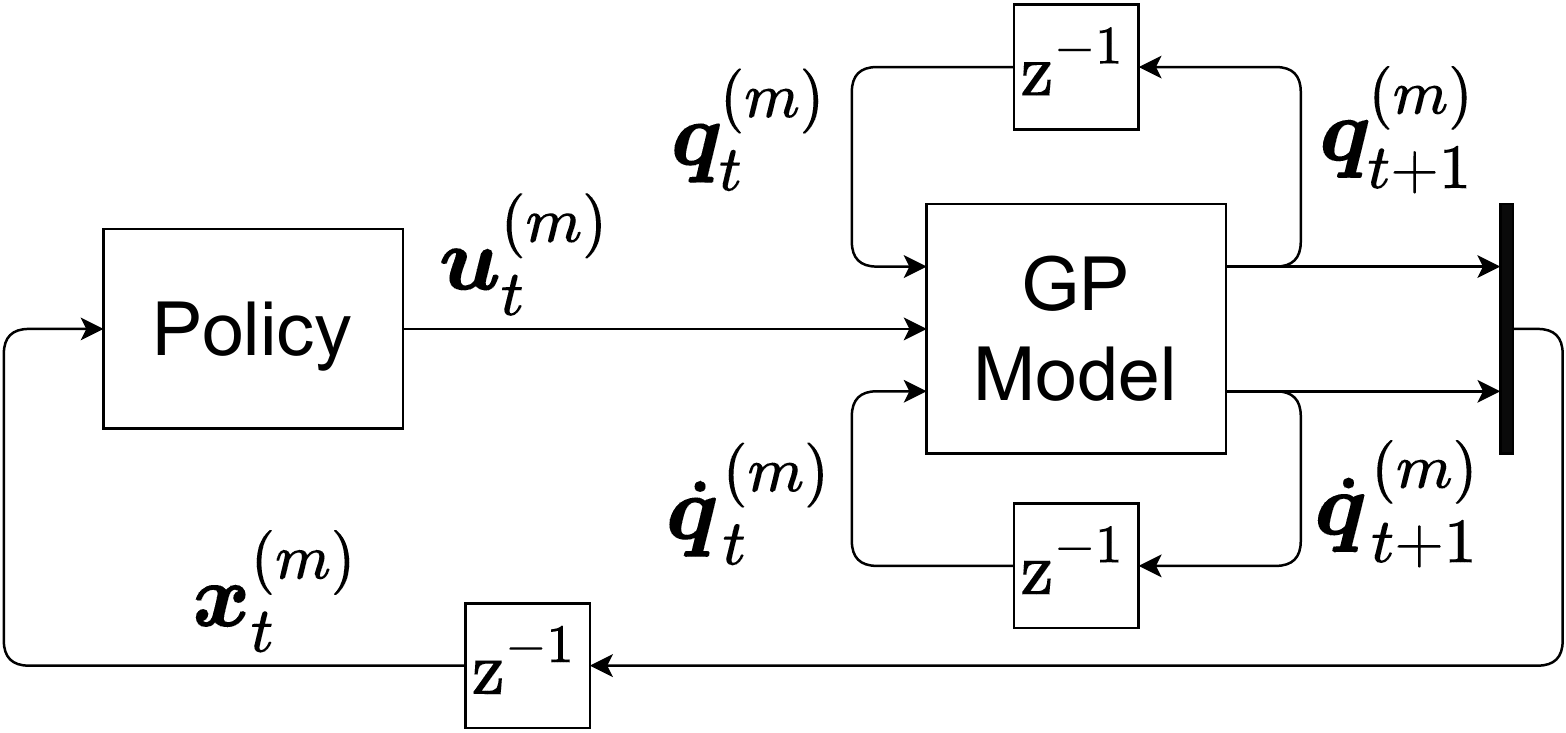}
         \caption{MC-PILCO}
     \end{subfigure}
     \begin{subfigure}[t]{0.95\linewidth}
         \centering
         \includegraphics[width=\textwidth]{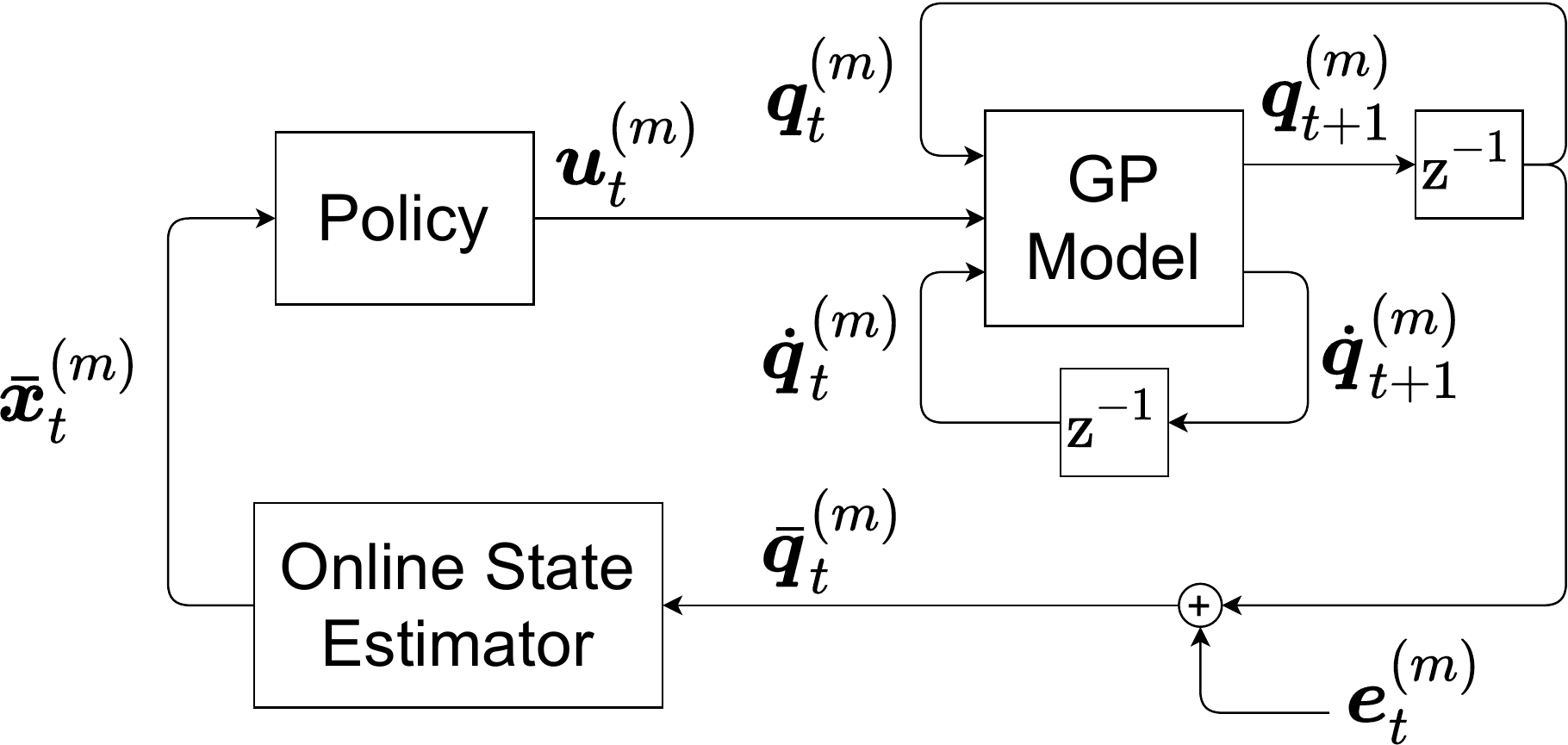}
         \caption{MC-PILCO4PMS}
     \end{subfigure}
        \caption{Block schemes illustrating particles generation in MC-PILCO (top) and MC-PILCO4PMS (bottom).}
        \label{fig:MCPILCO4PMS_scheme}
\end{figure}

\textbf{Simulation of the online estimators.} During policy optimization, instead of simulating only the evolution of the particles states, we simulate also the measurement system and the online estimators. The state fed to the policy, denoted by $\bar{\boldsymbol{x}}_t$, is computed to resemble the state that will be estimated online. Given the \emph{m}-th particle, this is given by
\begin{align*}
    \bar{\boldsymbol{x}}^{(m)}_t = \varphi\left(\bar{\boldsymbol{q}}^{(m)}_t\dots \bar{\boldsymbol{q}}^{(m)}_{t-m_q},\bar{\boldsymbol{x}}^{(m)}_{t-1}\dots \bar{\boldsymbol{x}}^{(m)}_{t-1-m_{\varphi}}\right) \text{,}
\end{align*}
where $\varphi$ denotes the online state estimator, with memory $m_q$ and $m_{\varphi}$, and $\bar{\boldsymbol{q}}^{(m)}_t$ is a fictitious noisy measurement of the \emph{m}-th particle positions. More precisely, let $\boldsymbol{q}^{(m)}_t$ the positions of the $\boldsymbol{x}^{(m)}_t$ particle state, then, we have
\begin{equation}\label{eq:particles noise}
    \bar{\boldsymbol{q}}^{(m)}_t = \boldsymbol{q}^{(m)}_t + \boldsymbol{e}^{(m)}_t\text{,}
\end{equation}
where $\boldsymbol{e}^{(m)}_t \in \mathbb{R}^{d_x/2}$ is Gaussian i.i.d. noise with zero mean and covariance $\text{diag}([\sigma^{(1)}_{\bar{x}}\dots \sigma^{(d_x/2)}_{\bar{x}}])$. The $\sigma^{(i)}_{\bar{x}}$s values must be tuned in accordance with the properties of the measurement system, e.g., the accuracy of the encoder. Then, the control input of the \emph{m}-th particle is computed as $\pi_{\boldsymbol{\theta}}(\bar{\boldsymbol{x}}^{(m)}_t)$, instead of $\pi_{\boldsymbol{\theta}}(\boldsymbol{x}^{(m)}_t)$. Differences in particles generation between MC-PILCO and MC-PILCO4PMS are summed up in the block scheme reported in Figure \ref{fig:MCPILCO4PMS_scheme}.

\section{MC-PILCO: Ablation Studies}\label{sec:ablation}
In this section, we analyze several aspects affecting the performance of MC-PILCO, such as the shape of the cost function, the use of dropout, the kernel choice, and the probabilistic model adopted, namely, \textit{full-state} or \textit{speed-integration} dynamical model. The purpose of the analysis is to validate the choices made in the proposed algorithm, and show the effect that they have on the control learning procedure. MC-PILCO has been implemented in Python, exploiting the PyTorch library \cite{paszke2017pytorch} automatic differentiation functionalities\footnote{Code available at \url{https://www.merl.com/research/license/MC-PILCO}}.

We considered the swing-up of a simulated cart-pole, a classical benchmark problem, to perform the ablation studies. The system and the experiments are described in the following. The physical properties of the system are the same as the system used in PILCO \cite{deisenroth2011pilco}: the masses of both cart and pole are 0.5 [kg], the length of the pole is $L=0.5$ [m], and the coefficient of friction between cart and ground is 0.1. The state at each time step $t$ is defined as $\boldsymbol{x}_t=[p_t, \dot{p}_t, \theta_t, \dot{\theta}_t]$, where $p_t$ represents the position of the cart and $\theta_t$ the angle of the pole. The target states corresponding to the swing-up of the pendulum is given by $p^{des}=0$ [m] and $\vert \theta^{des} \vert = \pi$ [rad]. The downward stable equilibrium point is defined at $\theta_t = 0$ [rad]. As done in \cite{deisenroth2011pilco}, in order to avoid singularities due to the angles, $\boldsymbol{x}_t$ is replaced in the algorithm with the state representation
\begin{equation}
\boldsymbol{x}^*_t = [p_t, \dot{p}_t, \dot{\theta}_t , sin(\theta_t), cos(\theta_t)]
    \label{eq:x_star_cartpole}
\end{equation} The control action is the force that pushes the cart horizontally. In all following experiments, we considered white measurement noise with standard deviation of $10^{-2}$, and as initial state distribution $\mathcal{N}([0,0,0,0],\text{diag}([10^{-4},10^{-4},10^{-4},10^{-4}]))$. The sampling time is $0.05$ seconds. The policy is a \textit{squashed-RBF-network} with $n_b=200$ basis functions. It receives as input $\boldsymbol{x}^*_t$ and $u_{max}=10$ [N]. The exploration trajectory is obtained by applying at each time step $t$ a random control action sampled from $\mathcal{U}(-10,10)$. GP reduction techniques were not adopted.

In this work, in all the experiments carried out with MC-PILCO, the cost function is a saturating function with the same general structure. The saturation is given by a negative exponential of the $\boldsymbol{x}_t-\boldsymbol{x}^{des}$ squared norm, namely,
\begin{equation*}
    c(\boldsymbol{x}_t) = 1 - \text{exp}\left(-\left(\boldsymbol{x}_t-\boldsymbol{x}^{des}\right)^T L\left(\boldsymbol{x}_t-\boldsymbol{x}^{des}\right)\right),
\end{equation*}
where $L$ is a diagonal matrix. The diagonal elements of $L$ are the inverse of the squared cost length-scales, and they allow weighting the different components of $\boldsymbol{x}_t-\boldsymbol{x}^{des}$, for instance based on their range of variation. Notice that this general structure of the cost can be applied to any system, and generalizes also to tasks with time-variant target, such as trajectory tracking tasks. Then, the cost function considered for the cart-pole cost is the following,
\begin{equation}\label{eq:abs_cost_cartpole}
    c(\boldsymbol{x}_t) = 1-\text{exp}\left(-\left(\frac{|\theta_t|-\pi}{l_{\theta}}\right)^2 -\left(\frac{p_t}{l_p}\right)^2\right),
\end{equation}
where the absolute value on $\theta_t$ is needed to allow different swing-up solutions to both the equivalent target angles of the pole, $\pi$ and $-\pi$.  The length-scales $l_{\theta}$ and $l_p$ define the shape of the cost function as $c(\cdot)$ goes to its maximum value more rapidly with small length-scales. Therefore, higher cost is associated to the same distance from the target state with lower $l_{\theta}$ and $l_p$. The lower the length-scale the more selective the cost function.

Other algorithms, like PILCO \cite{deisenroth2011pilco} and Black-DROPS \cite{chatzilygeroudis2017black}, used an alternative cost function for solving the cart-pole swing-up, with the saturation given by the negative exponential of the squared Euclidean distance between $\boldsymbol{x}_t$ and $\boldsymbol{x}^{des}$, namely,
\begin{equation}\label{eq:pilco_cost}
c^{\text{pilco}}(\boldsymbol{x}_t) = 1- \text{exp}\left(-\frac{1}{2} \left(\frac{d_t}{0.25}\right)^2 \right),
\end{equation}
where $d_t^2=p_t^2 + 2 p_t L sin(\theta_t) + 2 L^2 (1+ cos(\theta_t))$ is the squared euclidean distance between the tip of the pole and its position at the unstable equilibrium point with $p_t=0$ [m]. Since we compare MC-PILCO with PILCO and Black-DROPS in Section \ref{sec:comparison}, the results for the cart-pole system are rendered w.r.t. \eqref{eq:pilco_cost} to allow direct comparisons with previous literature.

All the comparisons consist of a Monte-Carlo study composed of 50 experiments. Every experiment is composed of 5 trials, each of length 3 seconds. The random seed varies at each experiment, corresponding to different explorations and initialization of the policy, as well as different measurement noise realizations.
For each trial, we report the median value and confidence interval defined by the 5-th and 95-th percentile of the cumulative cost computed with $c^{\text{pilco}}(\cdot)$, as well as the success rates observed. We mark two values of the cumulative cost indicatively associated with a swing-up for which the pole oscillates once or twice before reaching the upwards equilibrium. Trivially, the solution we aim for is the one that entails only one oscillation. Finally, we label a trial as "success" if $|p_t|< 0.1$ [m] and $170 \text{ [deg]} < |\theta_t| < 190 \text{ [deg]}$ $ \forall t$ in the last second of the trial.

To evaluate the statistical significance of the reported results, we tested the cumulative cost distributions with a  Mann-Whitney U-test \cite{Utest}, and the success rates with a Barnard's exact test \cite{Barnard1945ANT}. The significance level of both tests is set to~0.05. For the sake of space, we point out statistically significant results on the plots and tables and we explicitly report p-values only when objective conclusions are drawn.

\subsection{Cost shaping}\label{sec:cost_shaping}
\begin{figure}[t]
\centering
  \includegraphics[width=\linewidth]{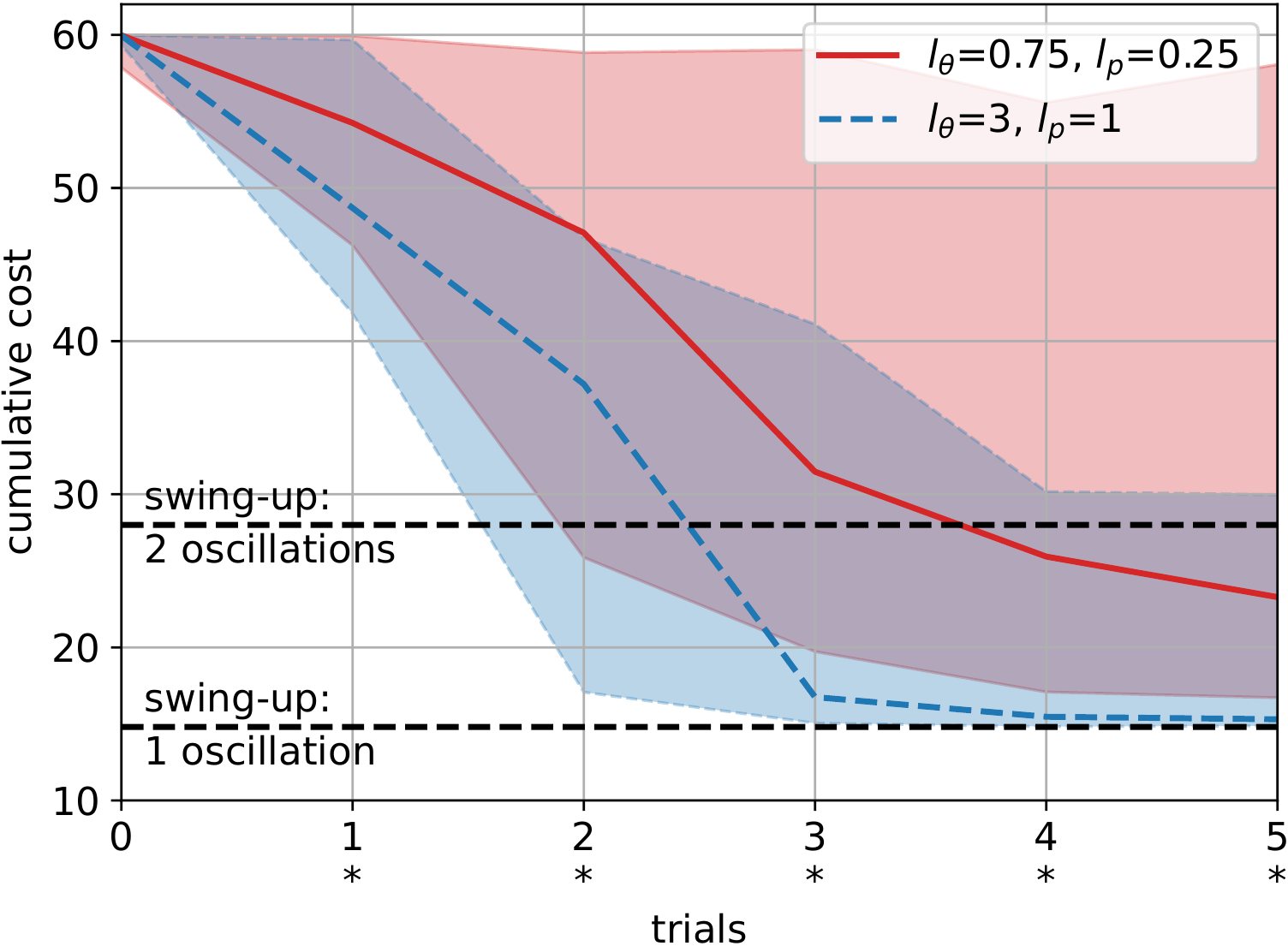}
  \caption{\small Median and confidence intervals of the cumulative cost $c^{\text{pilco}}(\cdot)$ per trial obtained using $(l_{\theta}=3,l_p=1)$ or $(l_{\theta}=0.75,l_p=0.25)$. In both cases, we used GP \textit{speed-integration} models with SE kernels and no dropout was applied. In the cumulative cost plot, we marked each trial with an *, to indicate the statistical significance of the difference between the two options. Instead, the difference between success rates is not statistically significant.
  }
  \smallskip
  \begin{tabular}{|l|l|l|l|l|l|}
  \multicolumn{6}{c}{Success Rates}\\
  \hline
   & \small{Trial 1} & \small{Trial 2} & \small{Trial 3} & \small{Trial 4} & \small{Trial 5} \\ \hline
  \small{$l$=(0.75,0.25)}       & \small{0\%} &\small{ 4\%}  & \small{42\%} & \small{68\%} & \small{70\%} \\ \hline
  \small{$l$=(3,1)} & \small{0\%} & \small{6\%} & \small{54\%} & \small{72\%} & \small{82\%} \\ \hline
  \end{tabular}

  \label{fig:confronto_costo}
\end{figure}

The first test regards the performance obtained varying the length-scales of the cost function in \eqref{eq:abs_cost_cartpole}. Reward shaping is a known important aspect of RL and here we analyze it for MC-PILCO. In Figure \ref{fig:confronto_costo}, we compare the evolution of the cumulative costs obtained with $(l_{\theta}=3,l_p=1)$ and $(l_{\theta}=0.75,l_p=0.25)$ and we report the observed success rates. The latter set of length-scales defines a more selective cost as the function shape becomes more skewed. In both cases, we adopted the \textit{speed-integration} model with SE kernel and no dropout was used during policy optimization.
The results show that with $(l_{\theta}=3,l_p=1)$ MC-PILCO performs better. Indeed, the median and variance of $(l_{\theta}=0.75,l_p=0.25)$ are higher  w.r.t. the ones of $(l_{\theta}=3,l_p=1)$ (the difference is statistically relevant at every trial, with p-value $2.7\cdot 10^{-4}$ at trial 1 and smaller than $10^{-4}$ in all subsequent trials). Observing the cumulative costs, it is possible to appreciate also a difference in the quality of the policies learned in the two cases. When using $(l_{\theta}=3,l_p=1)$, MC-PILCO learned to swing-up the cart-pole with only one oscillation in the majority of the experiments, while it has never been obtained with $(l_{\theta}=0.75,l_p=0.25)$. The success rates obtained with $(l_{\theta}=3,l_p=1)$ are greater than the counterpart, but this difference is not statistically significant, showing that the benefits of less selective cost functions are not sufficient, alone, to guarantee a clear advantage in terms of success rates.


These facts suggest that the use of too selective cost functions might decrease significantly the probability of converging to a solution. The reason might be that with small valued length-scales, $c(\boldsymbol{x}_t)$ is very peaked, resulting in almost null gradient, when the policy parameters are far from a good configuration, and increasing the probability of getting stuck in a local minimum. Instead, higher values of the length-scales promote the presence of non-null gradients also far away from the objective, facilitating the policy optimization procedure. These observations have already been made in PILCO, but the authors did not encountered difficulties in using a small length-scale such as 0.25 in (\ref{eq:pilco_cost}). This may be due to the analytic computation of the policy gradient made possible thanks to moment matching, as well as to the different optimization algorithm used. On the other hand, the length-scales' values seems to have no effect on the precision of the learned solution. To confirm this, in Table \ref{tab: precision} (rows 4 and 5), are reported the average distances from the target states obtained by successful policies at trial 5 during the last second of interaction. No significant difference in terms of precision in reaching the targets is observed.

\subsection{Dropout}\label{sec: ablation dropout}
\begin{figure}[t]
\centering
  \includegraphics[width=\linewidth]{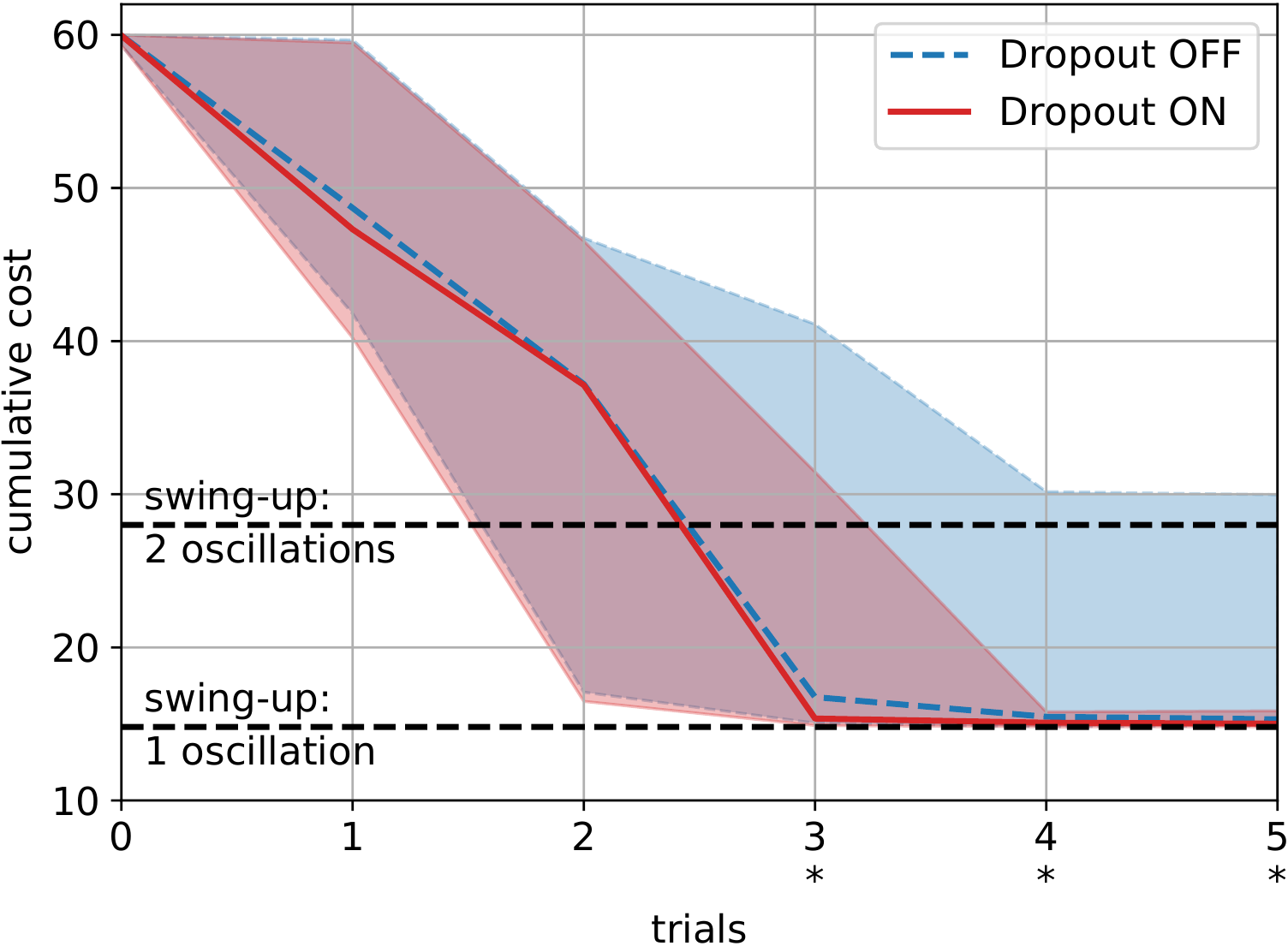}
  \caption{\small Median and confidence intervals of the cumulative cost $c^{\text{pilco}}(\cdot)$ per trial obtained using, or not, dropout. In both cases, we adopted GP \textit{speed-integration} model  with SE kernels, $l_{\theta}=3$ and $l_p=1$. Success rates are reported below. In both cumulative cost plot and success rate table, we marked each trial with an *, to indicate the statistical significance of the difference between the two options.}
  \smallskip
  \begin{tabular}{|l|l|l|l|l|l|}
  \multicolumn{6}{c}{Success Rates}\\
  \hline
    & \small{Trial 1} & \small{Trial 2} & \small{Trial 3} & \small{Trial 4} & \small{Trial 5} \\ \hline
  \small{Dropout OFF} & \small{0\%} & \small{6\%} & \small{54\%*} & \small{72\%*} & \small{82\%*} \\ \hline
  \small{Dropout ON}  & \small{0\%} & \small{14\%}  & \small{76\%*} & \small{98\%*} & \small{100\%*} \\ \hline
  \end{tabular}
  \label{fig:confronto_dropout}
\end{figure}

In this test, we compared the results obtained using, or not, the dropout during policy optimization. In Figure \ref{fig:confronto_dropout}, we compare the evolution of the cumulative cost obtained in the two cases and we show the obtained success rates.
In both scenarios, we adopted the \textit{speed-integration} model with SE kernel and a cost function with length-scales $(l_{\theta}=3,l_p=1)$. When using dropout, MC-PILCO solved the task at trial 4 in the 98\% of the experiments, and it managed to reach a 100\% success rate by trial 5. Instead, without dropout, the correct policy was not always found, even in the last trial. Notice that, when dropout is not used, the upper bounds of the cumulative costs in the last two trials are higher, meaning that the task cannot always be solved correctly. The statistical tests show that the advantages of dropout are statistically significant from trial 3 to trial 5 (cumulative cost p-values$: [0.33, 1.1, 0.29]\cdot 10^{-3}$; success rate p-values$: [11, 0.13, 0.90] \cdot 10^{-3}$). This fact suggests that dropout increases the probability of escaping from local minima, promoting the identification of a better policy. Additionally, Table \ref{tab: precision} (rows 3 and 5), shows that dropout also helps in decreasing the cart positioning error at the end of the swing-up (in both mean and standard deviation). Thus, we found empirically that dropout not only helps in stabilizing the learning process and in finding better solutions more consistently, but it can also improve the precision of the learned policies.

\subsection{Kernel function}

\begin{figure}[t]
\centering
  \includegraphics[width=\linewidth]{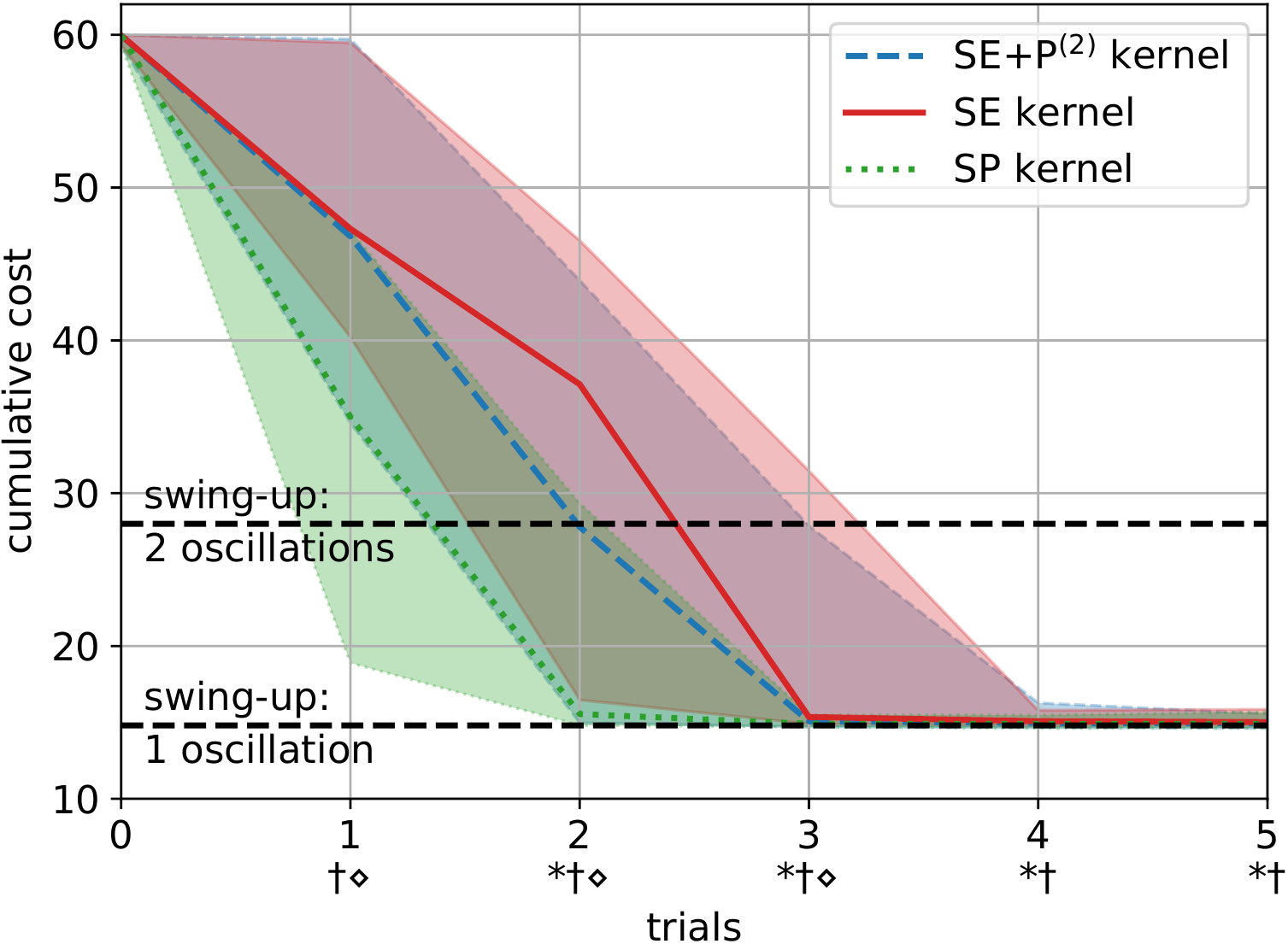}
  \caption{\small Median and confidence intervals of the cumulative cost $c^{\text{pilco}}(\cdot)$ per trial obtained using GP \textit{speed-integration} model with kernel SE, SE+P$^{(2)}$ and SP. In all the cases, $l_{\theta}=3$, $l_p=1$, and dropout was used. Success rates are reported below. In both cumulative cost plot and success rate table, we marked each trial to indicate the statistical significance of the difference between the three options. The labels adopted are,  *: SE+P$^{(2)}$ vs SE; $\dag$: SP vs SE; $\diamond$: SP vs SE+P$^{(2)}$.}
  \smallskip
  \begin{tabular}{|l|l|l|l|l|l|}
  \multicolumn{6}{c}{Success Rates}\\
  \hline
  & \small{Trial 1} & \small{Trial 2} & \small{Trial 3} & \small{Trial 4} & \small{Trial 5} \\ \hline
  \small{SE}                & \small{0\%$\dag$} & \small{14\%*$\dag$} & \small{76\%$\dag$} & \small{98\%} & \small{100\%} \\ \hline
  \small{SE+P$^\text{(2)}$} & \small{0\%$\diamond$} & \small{36\%*$\diamond$} & \small{88\%$\diamond$} & \small{98\%} & \small{100\%} \\ \hline
  \small{SP} & \small{8\%$\dag\diamond$} & \small{70\%$\dag\diamond$} & \small{100\%$\dag\diamond$} & \small{100\%} & \small{100\%} \\ \hline
  \end{tabular}
  \label{fig:confronto_kernel}
\end{figure}

In this test, we compared the results obtained using as kernels the SE, the SE+P$^{(2)}$ or the SP, see Section~\ref{sec:modelLearning}. Our aim is to test if the use of structured kernels can increase data efficiency. The kernels are listed from the least to the most structured: SE+P$^{(2)}$ can capture polynomial contributions more efficiently than SE, which are typical of robotic systems, and the SP kernel favours modes derived from the system equations (without assuming to know physical parameters)\footnote{SP basis functions are obtained by isolating, in each ODE defining cart-pole laws of motion, all the state-dependent components that are linearly related. In particular, we have $
\phi_{\dot{p}} (\boldsymbol{x}, u) = [\dot{\theta}^2\;sin(\theta), sin(\theta)cos(\theta), u, \dot{x}] $
 for the cart velocity GP, and $
\phi_{\dot{\theta}} (\boldsymbol{x}, u) = [\dot{\theta}^2\;sin(\theta)cos(\theta), sin(\theta), u\;cos(\theta), \dot{x}\;cos(\theta)] $ for the pole velocity GP.}. In all the cases, we adopted a \textit{speed-integration} model, the cost function was defined with length-scales $(l_{\theta}=3,l_p=1)$, and dropout was used. In Figure~\ref{fig:confronto_kernel}, we present, for each trial, the obtained cumulative costs and success rates. We can observe that the use of structured kernels, such as SP and SE+P$^{(2)}$, can be beneficial in terms of data efficiency, compared to adopting the standard SE kernel. In fact,  the fastest convergence is observed in the SP case, where a success rate of 100\% is obtained at trial 3, after only 9 seconds of experience. Also at trial 2, the gap between the SP performance and the ones of SE and SE+P$^{(2)}$ is considerable. The statistical tests show that the differences w.r.t the SE+P$^{(2)}$ and SE kernel are statistically significant from trial 1 to trial 3, confirming the augmented data efficiency (SP vs SE+P$^{(2)}$ cumulative cost p-values:  $<10^{-4}$ at trials 1 and 2, $3.9\cdot 10^{-3}$ at trial 3; SP vs SE+P$^{(2)}$ success rate p-values$: [22, 0.37, 6.0]\cdot 10^{-3}$ ; SP vs SE cumulative cost p-values: always $<10^{-4}$;  SP vs SE success rate p-values: $2.2\cdot 10^{-2}$ at trial 1 and $<10^{-4}$ later). Moreover, the cumulative cost distributions obtained by SE+P$^{(2)}$ and SE differ statistically after trial 1 (p-values: $<10^{-4}$ at trial 2, $[0.42, 3.6, 6.2]\cdot10^{-3}$ later), observing a statistically significant success rate improvement at trial 2 (p-value$: 6.0\cdot10^{-3}$) when comparing the performance of SE+P$^{(2)}$ and SE kernels. These differences can be explained by the capacity of a more structured kernel to better generalize outside of the training set, i.e., to learn dynamical properties of the system that hold also in areas of the state-action space with scarce data points. In fact, some dynamics components of the cart-pole system are polynomial functions of the GP input $\tilde{\boldsymbol{x}}_t = (\boldsymbol{x}^*_t, \boldsymbol{u}_t)$, with $\boldsymbol{x}^*_t$ defined in \eqref{eq:x_star_cartpole}, leading SE+P$^{(2)}$ to achieve better data efficiency during the first trials compared to SE. With one step further, the SP kernel exploits features determined by a direct knowledge of the physical model, thus it reaches a even higher level of data efficiency.

\subsection{Speed-integration model}
In this test, we compared the performance obtained  by the proposed \textit{speed-integration} dynamical model and  by the standard \textit{full-state} model. In both cases, SE kernels were adopted, the cost function was defined with length-scales $(l_{\theta}=3,l_p=1)$, and dropout was used. The success rates obtained at each trial are reported in Table \ref{tab:SIvsFS_model}. We can observe that the performance obtained by the two structures are quite similar, in fact the differences between success rates observed at trial 2 and 3 are not statistically significant. Also the precision in reaching the target state is comparable, as reported in Table \ref{tab: precision} (rows 3 and 6). Hence, the proposed \textit{speed-integration} model performs similarly compared to the \textit{full-state} counterpart but offers the advantage of reducing the computational burden by halving the number of GPs employed.

\begin{table}[h]
    \centering
    \begin{tabular}{|l|l|l|l|l|l|}
  \hline
   & \small{Trial 1} & \small{Trial 2} & \small{Trial 3} & \small{Trial 4} & \small{Trial 5} \\ \hline
  \small{Full-state} & \small{0\%} & \small{12\%} & \small{70\%} & \small{98\%} & \small{100\%} \\ \hline
  \small{Speed-int.} & \small{0\%} & \small{14\%} & \small{76\%} & \small{98\%} & \small{100\%} \\ \hline
  \end{tabular}
  \caption{Success rates per trial obtained using \textit{full-state} or \textit{speed-integration} dynamical models. The difference between the two options is not statistically significant.}
    \label{tab:SIvsFS_model}
\end{table}

\section{MC-PILCO Experiments}\label{sec:expeirmentSim}
In this section, we describe different experiments conducted on simulated scenarios to test the validity of the proposed MC-PILCO algorithm. First, we compare MC-PILCO to other GP-based MBRL algorithms, namely PILCO and Black-DROPS, on the cart-pole benchmark. Second, we analyse MC-PILCO and PILCO computational time requirements. Moreover, we tested the capacity of our algorithm to handle bimodal state distributions in the cart-pole benchmark. Finally, we tested MC-PILCO in a higher DoF system, namely a UR5 robotic manipulator, where we solved a trajectory tracking task.
\subsection{Comparison with other algorithms}\label{sec:comparison}
\begin{figure}
\centering
  \includegraphics[width=\linewidth]{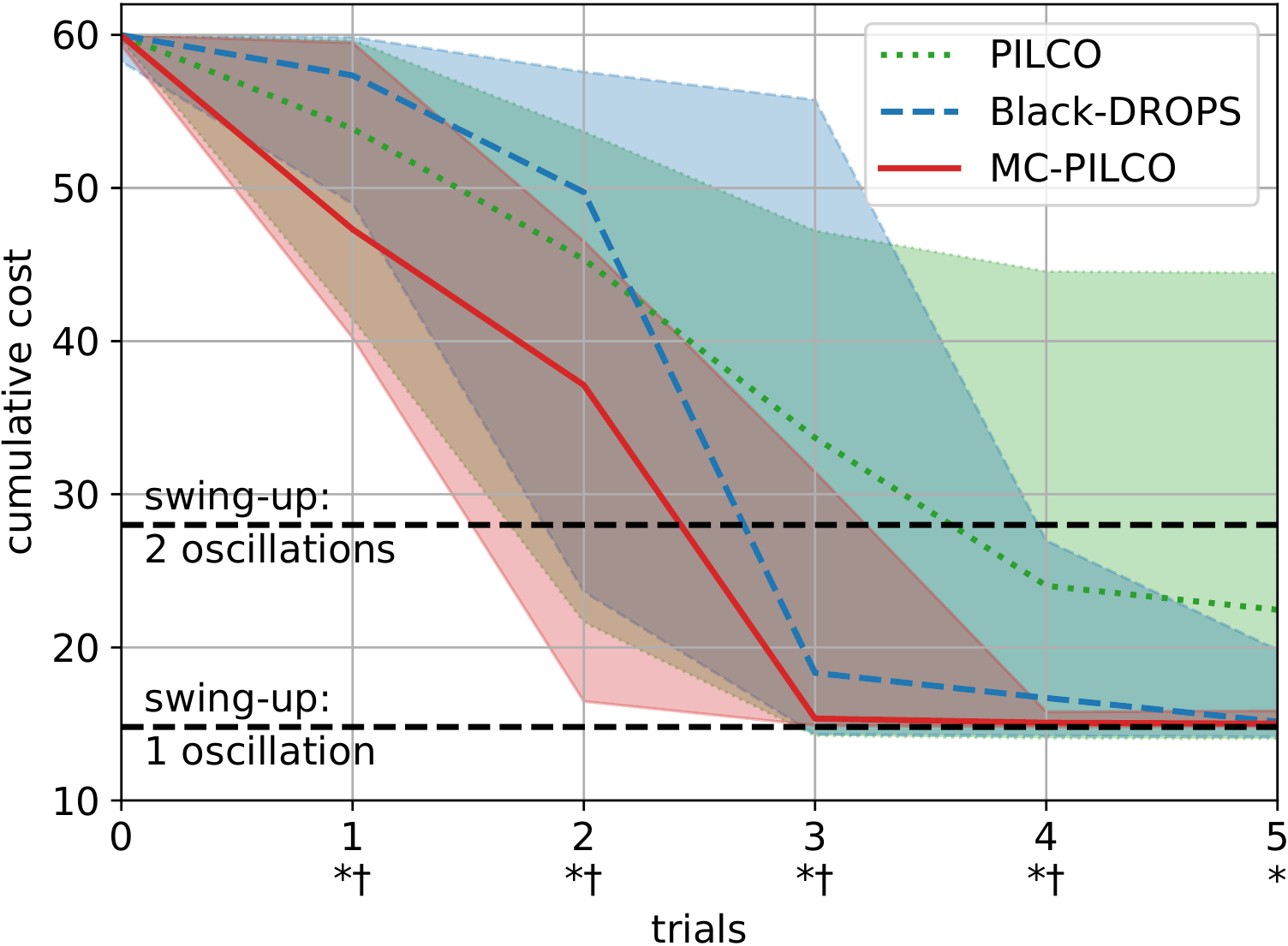}
  \caption{\small Median and confidence intervals of the cumulative cost $c^{\text{pilco}}(\cdot)$ per trial obtained with PILCO, Black-DROPS and MC-PILCO (with \textit{speed-integration} model, SE kernel, dropout activated, $l_{\theta}=3$ and $l_p=1$). Success rates are reported below. In both cumulative cost plot and success rate table, we marked each trial to indicate the statistical significance of the difference between the three algorithms. In the following, we report the list of labels adopted, *: MC-PILCO vs PILCO, $\dag$: MC-PILCO vs Black-DROPS}
  \smallskip
  \begin{tabular}{|l|l|l|l|l|l|}
  \multicolumn{6}{c}{Success Rates}\\
  \hline
  & \small{Trial 1} & \small{Trial 2} & \small{Trial 3} & \small{Trial 4} & \small{Trial 5} \\ \hline
  \small{PILCO} & \small{2\%} & \small{4\%*} & \small{20\%*} & \small{36\%*} & \small{42\%*} \\ \hline
  \small{Black-DROPS} & \small{0\%} & \small{4\%$\dag$} & \small{30\%$\dag$} & \small{68\%$\dag$} & \small{86\%$\dag$} \\ \hline
  \small{MC-PILCO} & \small{0\%} & \small{14\%*$\dag$} & \small{76\%*$\dag$} & \small{98\%*$\dag$} & \small{100\%*$\dag$} \\ \hline
 \end{tabular}
  \label{fig:confronto_algoritmi}
\end{figure}

We tested PILCO\footnote{PILCO code available at \url{http://mlg.eng.cam.ac.uk/pilco/}}, Black-DROPS\footnote{Black-DROPS code available at \url{https://github.com/resibots/blackdrops}}, and MC-PILCO on the cart-pole system, previously described in Section~\ref{sec:ablation}. In MC-PILCO, we considered the cost function \eqref{eq:abs_cost_cartpole} with length-scales $(l_{\theta}=3,l_p=1)$, and adopted the SE kernel, as it is the one employed by the other algorithms. PILCO and Black-DROPS optimized their original cost/reward function \eqref{eq:pilco_cost}. To be consistent with the previous literature, we used the latter cost function as common metric to compare the results. For fairness, we verified if also PILCO and Black-DROPS benefits from higher length-scales in \eqref{eq:pilco_cost}. Moreover, we tested Black-DROPS with cost function \eqref{eq:abs_cost_cartpole} and increasing the length-scales from small values to $(l_{\theta}=3,l_p=1)$. The performance of both the algorithms deteriorated as we increased the length-scales. For these reasons, we report the results of both algorithms achieved with \eqref{eq:pilco_cost}, which gave the best performance. The observed cumulative costs and success rates are reported in Figure \ref{fig:confronto_algoritmi}. MC-PILCO achieved the best performance both in transitory and at convergence. In fact, it obtained a statistically significant improvement in terms of success rate w.r.t. the other algorithms from trial 2 to 5 (MC-PILCO vs PILCO p-values: $4.7\cdot10^{-2}$ at trial 2 and $<10^{-4}$ later; MC-PILCO vs Black-DROPS p-values: $4.7\cdot10^{-2}$ at trial 2, $<10^{-4}$ at trials 3 and 4, and $3.3\cdot10^{-3}$ at trial 5). Moreover, MC-PILCO cumulative cost distributions show lower median and variance w.r.t. counterparts, with differences always statistically significant up to trial 4 (MC-PILCO vs PILCO, p-values: $3.5\cdot10^{-3}$ at trial 1 and $<10^{-4}$ later; MC-PILCO vs Black-DROPS p-values: $<10^{-4}$ at trial 1 and 2, $4.6\cdot10^{-4}$ at trial 3 and $1.1\cdot10^{-2}$ at trial 4). On the contrary, PILCO showed poor convergence properties, while Black-DROPS can outperform PILCO, but without reaching MC-PILCO level of performance. Finally, results in Table \ref{tab: precision} (rows 1, 2, 3, 7 and 8), also show that MC-PILCO policies are more precise in reaching the target.

\begin{table}[ht]
\centering
\begin{tabular}{|l|l|l|l|}
\hline
\multicolumn{2}{|l|}{ }                                     & $e_p$ [m]     & $e_{\theta}$ [rad] \\ \hline
1&S.I. SE+P$^{(2)}$ (3,1) drop. on & $0.008\pm0.003$ & $0.011\pm0.04$    \\ \hline
2&S.I. SP (3,1) drop. on           & $0.008\pm0.003$ & $0.011\pm0.005$    \\ \hline
3& S.I. SE (3,1) drop. on           & $0.010\pm0.005$ & $0.011\pm0.005$    \\ \hline
4& S.I. SE (0.75,0.25) drop. off    & $0.016\pm0.009$ & $0.012\pm0.008$    \\ \hline
5& S.I. SE (3,1) drop. off          & $0.019\pm0.014$ & $0.015\pm0.009$    \\ \hline
6& F.S. SE (3,1) drop. on           & $0.011\pm0.005$ & $0.011\pm0.005$    \\ \hline
7& Black-DROPS                      & $0.025\pm0.011$ & $0.033\pm0.019$    \\ \hline
8& PILCO                            & $0.027\pm0.012$ & $0.045\pm0.019$    \\ \hline

\end{tabular}
\caption{\small Average distances from the target states ($p_t=0$ and $\theta_t=\pm\pi$) obtained during the last second of interaction with the cart-pole by the successful policies learned by PILCO, Black-DROPS and the various MC-PILCO configurations analyzed in Section \ref{sec:ablation}. Different configurations are labeled reporting the adopted dynamical model structure (\textit{speed-integration}, S.I., or \textit{full-state}, F.S.), kernel function, cost length-scales, and if dropout was used or not. Values are reported as mean $\pm$ standard deviation, calculated over the total number of successful runs at trial 5.}
\label{tab: precision}
\end{table}

\subsection{Computational time analysis}
\begin{figure}
\centering
  \includegraphics[width=1\linewidth]{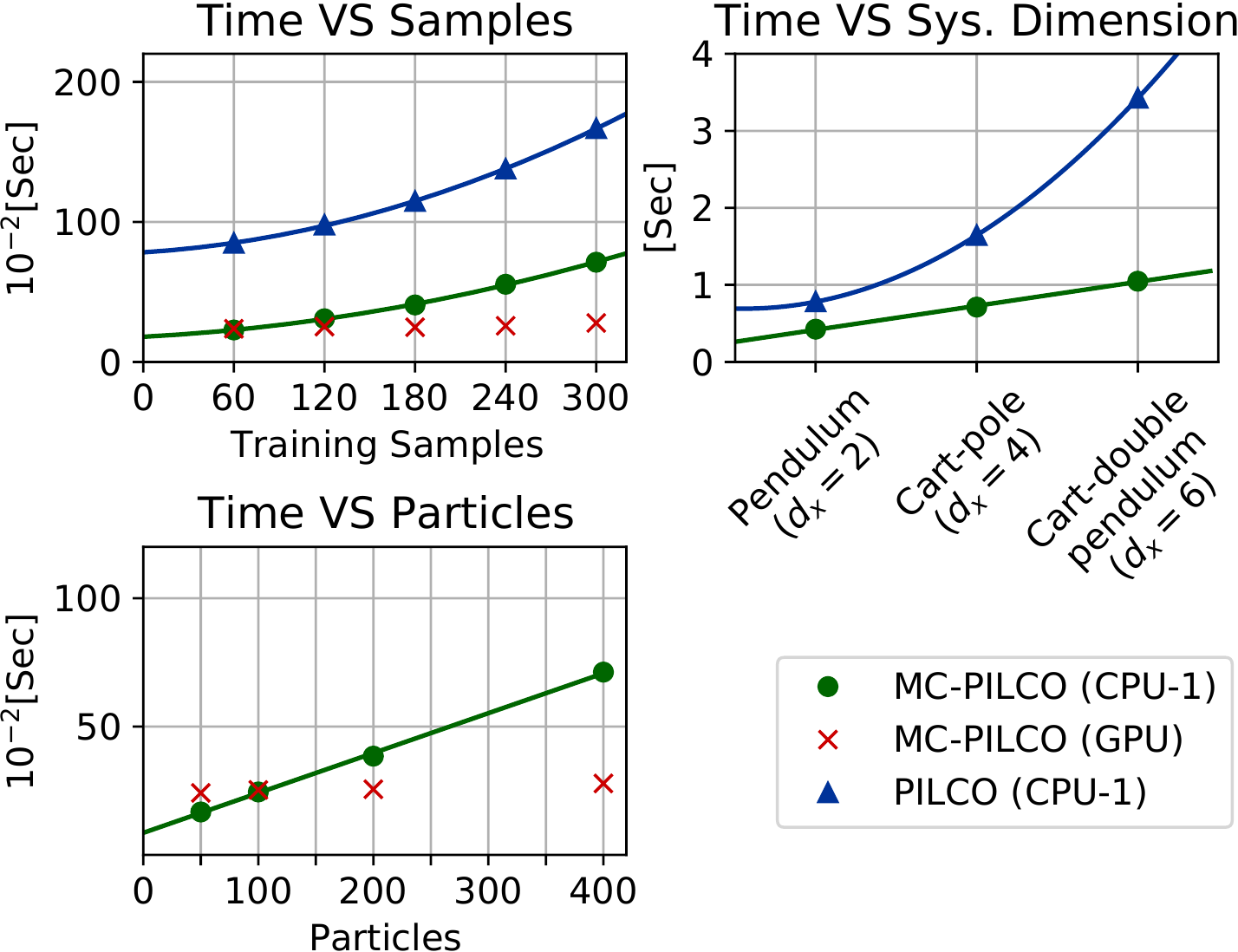}
  \caption{\small Average time required to compute the distribution of long-term predictions and its gradient as a function of: GP training samples (top-left, on the simulated cart-pole), system dimension (top-right, with 300 training samples), number of particles (bottom-left, with 300 training samples on the simulated cart-pole). For all the algorithms and systems, the policy was a RBF network with 200 basis functions. Hardware adopted: CPU: Intel i7-6700K, GPU: Nvidia RTX 2080 Ti.}
  \label{fig:computational_time}
\end{figure}

We analyzed the time required by MC-PILCO and PILCO to compute the approximation of the cumulative cost expectation and its gradient w.r.t. the policy parameters. We left Black-DROPS out of this comparison, because of the different nature of its optimization strategy, which is based on a black-box gradient-free algorithm. We remark that the algorithms are implemented in different languages, which significantly affects computational time (PILCO is implemented in MATLAB, MC-PILCO in Python). MC-PILCO relies on the \textit{speed-integration} dynamical model, which halves the number of GPs employed. For these reasons, we are more interested in the behavior of computational time as a function of training samples and system dimension than in absolute values of time reported.  Figure \ref{fig:computational_time} shows that both with MC-PILCO and PILCO the average computational time scales with the square of the training samples $n$, as expected from the analysis in Section \ref{subsec:longterm_prediction}. As regards the dependencies w.r.t. system dimensions, we considered three systems of increasing dimension: a pendulum ($d_x=2$), a cart-pole ($d_x=4$), and a cart-double-pendulum ($d_x=6$). MC-PILCO scales linearly, while for PILCO the linear model is not enough to fit the average computational time; PILCO scales at least quadratically. This fact represents a great advantage of the particles based approximation used by MC-PILCO w.r.t. the moment matching approach followed by PILCO. Figure \ref{fig:computational_time} also reports MC-PILCO computational time as a function of the particles number. In accordance with the results in Section \ref{subsec:longterm_prediction}, MC-PILCO  complexity scales linearly with the number of particles. Finally, we tested MC-PILCO on a GPU instead of a CPU: the average times collected are almost constant w.r.t. the number of samples and particles. As expected, MC-PILCO is highly parallelizable.

We conclude the computational time analysis reporting the average and the standard deviation of the time required to run MC-PILCO and PILCO for 5 trials, computed in the 50 runs. On average, PILCO and MC-PILCO took, respectively, $1692$ and $2060$[s], with standard deviations $94$ and $157$[s]. The times are similar, but PILCO is faster than MC-PILCO, even though it requires more time to compute a single approximation of the cumulative cost expectation and its gradient. This is due to the optimization algorithm adopted, which performs fewer steps but converges to worse policies. As previously highlighted, the performance gap between the two algorithms is considerable. At the last trial, PILCO converges only in $42\%$ of the runs, while MC-PILCO in $100\%$. For the sake of completeness, we tried to increase the maximum number of function evaluations admitted by the PILCO optimization algorithm. Computational time increased without improving success rate.

\subsection{Handling bimodal distributions}
\begin{figure}
\centering
  \includegraphics[width=1\linewidth]{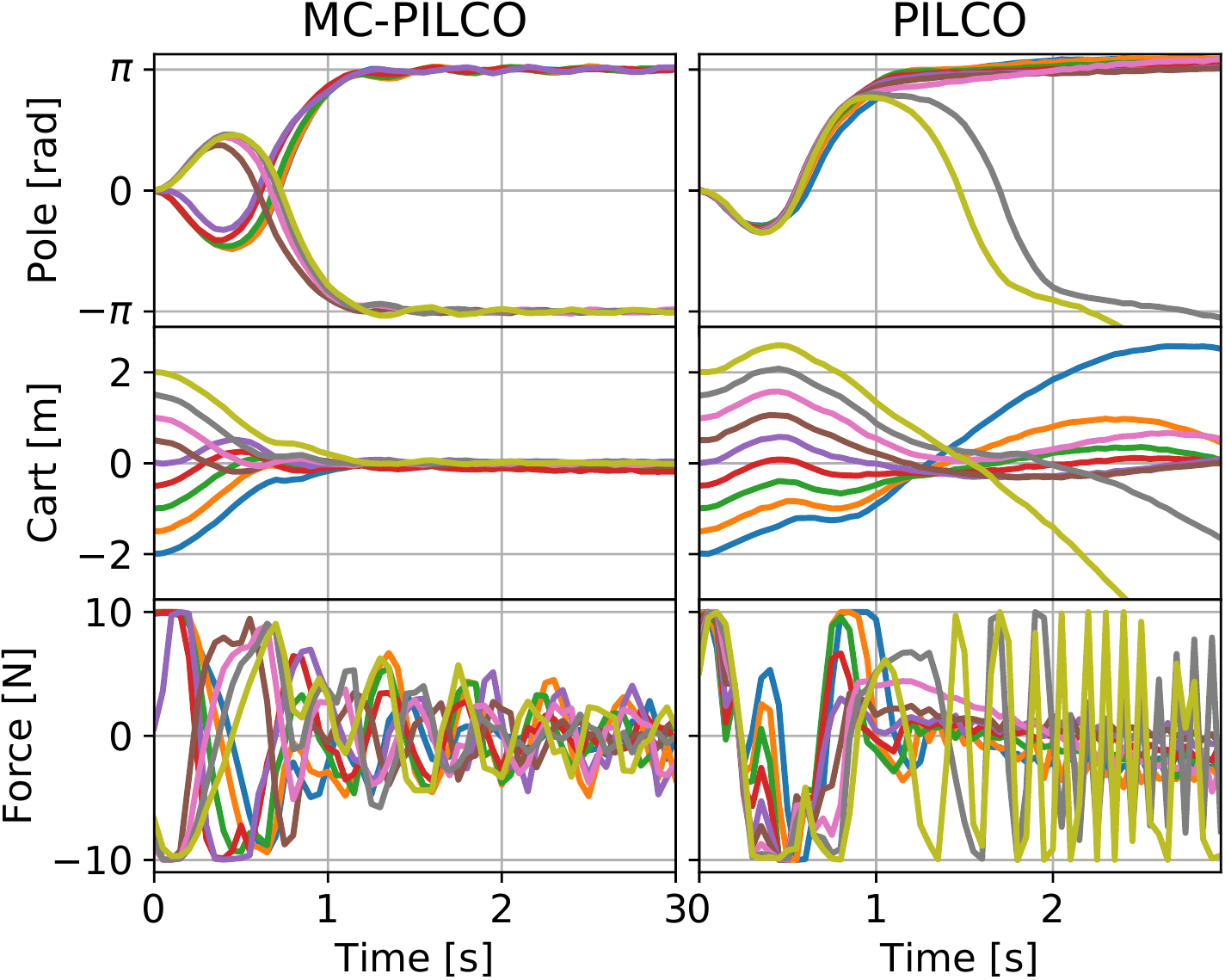}
  \caption{\small (Left) MC-PILCO policy applied to the cart-pole system starting from nine different sparse cart initial positions, namely: -2, -1.5, -1, -0.5, 0, 0.5, 1, 1.5, 2 [m], see middle figures and same pole angle. All 9 trajectories are reported in the figures.. The policy is able to complete the task in all cases, pushing the cart in different directions depending on its initial condition. The pole trajectories have a bimodal distribution. (Right) PILCO policy applied starting from the same cart initial positions. This policy struggles to adapt to different starting conditions, and it cannot swing up the cart-pole when starting from the initial positions further away from zero.}
  \label{fig:multimodal}
\end{figure}

One of the main advantages of particle-based policy optimization is the capability to handle multimodal state evolutions. This is not possible when applying methods based on moment matching, such as PILCO. We verified this advantage by applying both PILCO and MC-PILCO to the simulated cart-pole system, when considering a very high variance on the initial cart position, $\sigma_p^2 = 0.5$, which corresponds to have unknown cart's initial position (but limited within a reasonable range). With this initial condition, the optimal initial cart direction and the swing-up direction depend on whether the initial position of the cart is positive or negative. The aim is to be in a situation in which the policy has to solve the task regardless of the initial conditions and needs to have a bimodal behaviour in order to do so. Note that the situation described could be relevant in several real applications. We kept the same setup used in previous cart-pole experiments, changing the initial state distribution to a zero mean Gaussian with covariance matrix $\text{diag}([0.5,10^{-4},10^{-4},10^{-4}]))$. MC-PILCO optimizes the cost in \eqref{eq:abs_cost_cartpole} with length-scales $(l_{\theta}=3,l_p=1)$. We tested the policies learned by the two algorithms starting from nine different cart initial positions (-2, -1.5, -1, -0.5, 0, 0.5, 1, 1.5, 2 [m]). In Section \ref{sec:comparison}, we observed that~PILCO struggles to consistently converge to a solution and the high variance in the initial conditions accentuates this issue. Nevertheless, in order to make the comparison possible, we cherry-picked a random seed for which PILCO converged to a solution in this particular scenario. In Figure \ref{fig:multimodal}, we show the results of the experiment. MC-PILCO is able to handle the initial high variance. It learned a bimodal policy that pushes the cart in two opposite directions, depending on the cart's initial position, and stabilizes the system in all the experiments. On the contrary, PILCO's policy is not able to control the cart-pole for all the tested starting conditions. Its strategy is always to push the cart in the same direction, and it cannot stabilize the system when the cart starts far away from the zero position. The state evolution under MC-PILCO's policy is bimodal, while PILCO cannot find this type of solutions because of the unimodal approximation enforced by moment matching.

In this example, we have seen that a multimodal state evolution could be the correct solution, when starting from a unimodal state distribution with high variance, due to dependencies on initial conditions. In other cases, multimodality could be directly enforced by the presence of multiple possible initial conditions that would be badly modeled with a single unimodal distribution. MC-PILCO can handle all these situations thanks to its particle-based method for long-term predictions. Similar results were obtained when considering bimodal initial distributions. 

\subsection{Trajectory tracking task on UR5 manipulator}
The objective of this experiment is to test MC-PILCO in a more complex system with higher DoF. We used MC-PILCO to learn a joint-space controller for a UR5 robotic arm (6 DoF) simulated in MuJoCo \cite{todorov2012mujoco}. Let the state at time $t$ be $\boldsymbol{x}_t = [\boldsymbol{q}_t^T, \dot{\boldsymbol{q}}_t^T]^T$, where $\boldsymbol{q}_t$,$\dot{\boldsymbol{q}_t}$ $\in \mathbb{R}^6$ are joint angles and velocities, respectively. The objective for the policy $\pi_{\boldsymbol{\theta}}$ is to control the torques $\boldsymbol{\tau}_t$ in order to follow a desired trajectory $(\boldsymbol{q}^r_t, \dot{\boldsymbol{q}}^r_t)$ for $t=0,\dots ,T$. Let $\boldsymbol{e}_t = \boldsymbol{q}^r_t-\boldsymbol{q}_t, \dot{\boldsymbol{e}}_t=\dot{\boldsymbol{q}}^r_t-\dot{\boldsymbol{q}}_t$ be position and velocity errors at time $t$, respectively. The policy is a multi-output \textit{squashed-RBF-network} with $n_b=400$ Gaussian basis functions and $u_{max}=1$ [N$\cdot$m] for all the joints, that maps states and errors into torques, $\pi_{\boldsymbol{\theta}}: \boldsymbol{q}_t,\dot{\boldsymbol{q}}_t,\boldsymbol{e}_t,\dot{\boldsymbol{e}}_t \mapsto ~\boldsymbol{\tau}_t$. The control scheme is represented in Figure \ref{fig:UR5ctrl_scheme}.
\begin{figure}
\centering
  \includegraphics[width=0.95\linewidth]{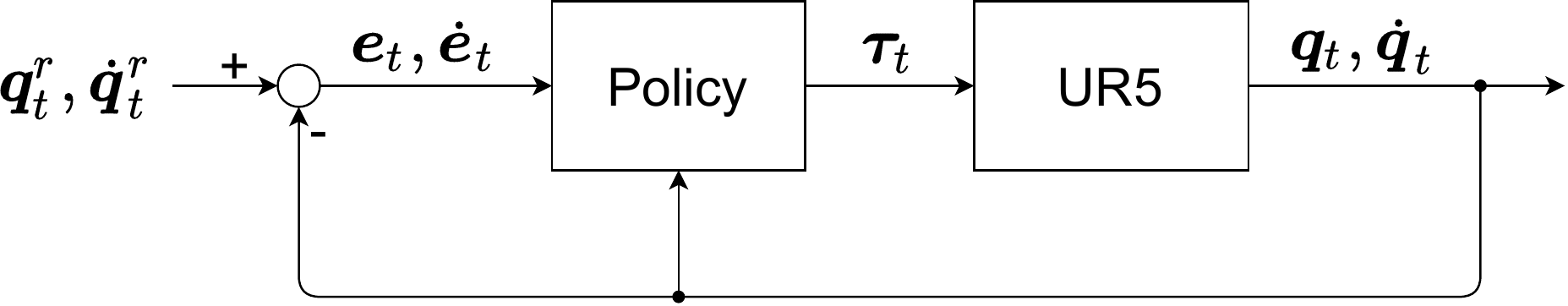}
  \caption{\small Joint-space control scheme for UR5 robotic arm.}
  \label{fig:UR5ctrl_scheme}
\end{figure}
\if
\tikzstyle{block} = [draw, fill=blue!15, rectangle,
    minimum height=2em, minimum width=4em]
\tikzstyle{sum} = [draw, fill=blue!15, circle, node distance=1.5cm]
\tikzstyle{input} = [coordinate]
\tikzstyle{output} = [coordinate]
\tikzstyle{pinstyle} = [pin edge={to-,thin,black}]
\begin{figure}[!h]
\begin{center}
    \begin{tikzpicture}[auto, node distance=2cm,>=latex']
        \node [input, name=input] {};
        \node [sum, right of=input] (sum) {};
        \node [block, right of=sum] (controller) {$\pi_{\boldsymbol{\theta}}$};
        \node [block, right of=controller, node distance=3cm] (system) {UR5};
        \draw [->] (controller) -- node[name=u] {$\boldsymbol{\tau}_t$} (system);
        \node [output, right of=system] (output) {};

        \draw [draw,->] (input) -- node {$\boldsymbol{q}^r_t, \dot{\boldsymbol{q}}^r_t$} (sum);
        \draw [->] (sum) -- node {$\boldsymbol{e}_t, \dot{\boldsymbol{e}}_t$} (controller);
        \draw [->] (system) -- node [name=y] {$\boldsymbol{q}_t, \dot{\boldsymbol{q}_t}$}(output);
        \draw [->] (y) -- ++ (0,-1.5) -| node [pos=0.99] {}
            node [near end] {$\boldsymbol{q}_t, \dot{\boldsymbol{q}}_t$} (controller);
        \draw [->] (y) -- ++ (0,-2) -| node [pos=0.99] {$-$}
            node [near end] {$\boldsymbol{q}_t, \dot{\boldsymbol{q}}_t$} (sum);
    \end{tikzpicture}
\end{center}
\caption{Joint-space control scheme for UR5 robotic arm.
}
\label{fig:UR5ctrl_scheme}
\end{figure}
\fi

In this experiment, we considered a control horizon of 4 seconds with a sampling time of 0.02 seconds. The reference trajectory has been calculated to make the end-effector draw a circle in the X-Y operational space. The initial exploration, used to initialize the \textit{speed-integration} dynamical model, is provided by a poorly-tuned PD controller. We used SE+$\text{P}^{\text{(1)}}$ kernels in the GP dynamical model. The GP reduction thresholds were set to $10^{-3}$. GP input was built using extended state $\boldsymbol{x}^*_t=[\dot{\boldsymbol{q}}_t, sin(\boldsymbol{q}_t), cos(\boldsymbol{q}_t)]$. $M=200$ is the number of particles used for gradient estimation. The cost function considered is defined as,
\begin{equation*}
    c(\boldsymbol{x}_t)=1-\text{exp}\left( -\left(\frac{||\boldsymbol{q}^r_t-\boldsymbol{q}_t||}{0.5}\right)^2  -\left(\frac{||\dot{\boldsymbol{q}}^r_t-\dot{\boldsymbol{q}}_t||}{1}\right)^2 \right).
\end{equation*}
We assumed full state observability with measurements perturbed by white noise with standard deviation of $10^{-3}$. The initial state distribution is a Gaussian centered on $(\boldsymbol{q}^r_0, \dot{\boldsymbol{q}}^r_0)$ with standard deviation of $10^{-3}$. Policy optimization parameters are the same reported in Table \ref{tab: standard_opt_setup}, with the exception of $n_s=400$ and $\sigma_s$ = 0.05, to enforce more restrictive exit conditions.

\begin{figure}[t]
\centering
  \includegraphics[width=0.9\linewidth]{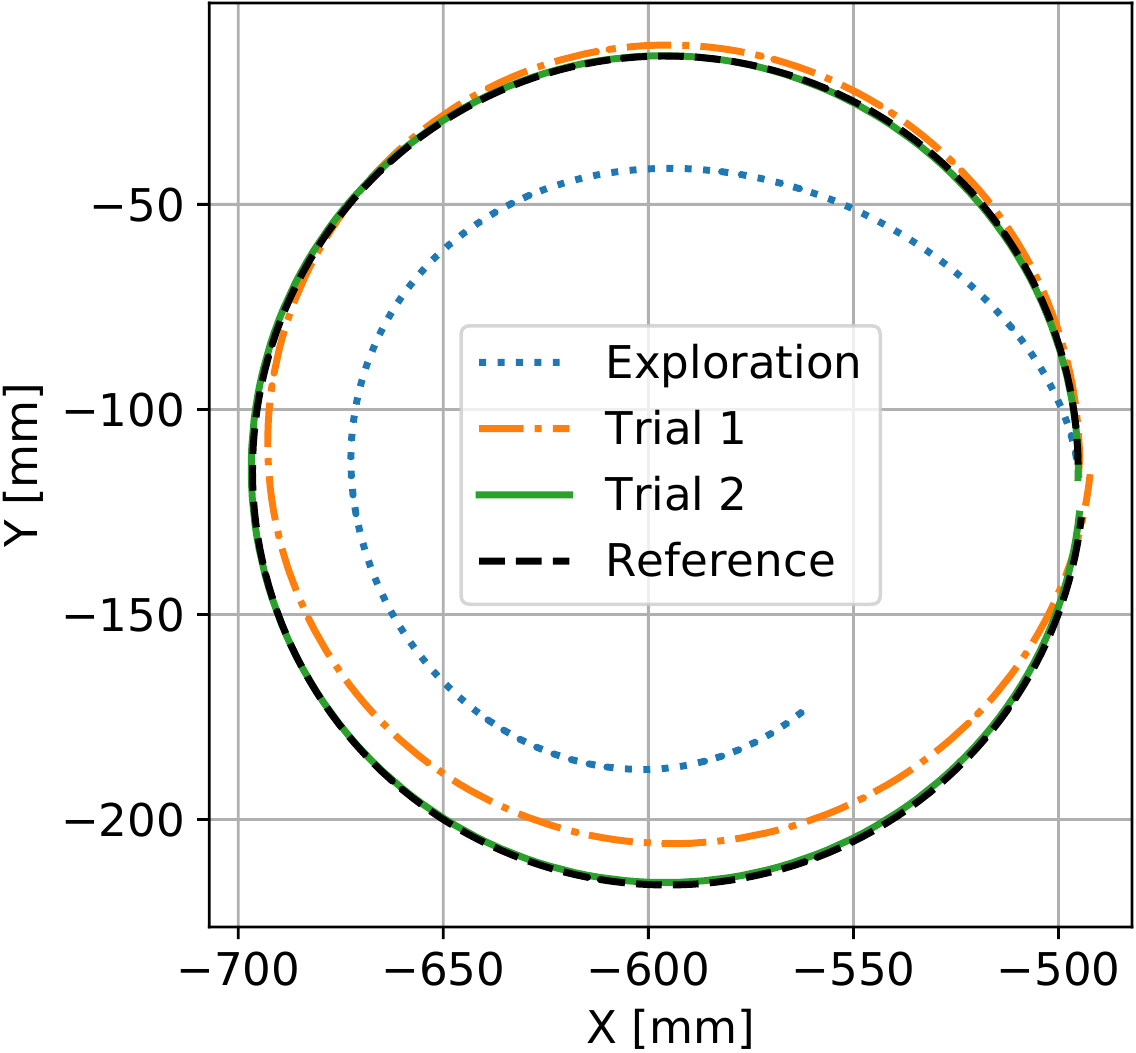}
  \caption{\small End-effector trajectories obtained in exploration and for each trial of policy learning together with the desired circle. Let $\boldsymbol{e}_{ee}$ be the error between the desired and the actual end-effector trajectories. In the table below, we report, in millimeters, the maximum and mean errors ($\pm$ 3$\times$standard deviation) at each trial.}
  \smallskip
  \begin{tabular}{|l|l|l|l|}
  \hline
   & \small{Exploration} & \small{Trial 1} & \small{Trial 2}  \\ \hline
  \small{mean($\boldsymbol{e}_{ee}$) [mm]}  & \small{140.66$\pm$158.94} & \small{21.15$\pm$41.71} & \small{0.65$\pm$0.69} \\ \hline
  \small{max($\boldsymbol{e}_{ee}$) [mm]}  & \small{196.70} & \small{40.79} & \small{1.08}  \\ \hline
\end{tabular}
  \label{fig:ur5end-effector}
\end{figure}

In Figure \ref{fig:ur5end-effector}, we report the trajectory followed by the end-effector at each trial, together with the desired trajectory. MC-PILCO considerably improved the high tracking error obtained with the PD controller after only 2 trials (corresponding to 8 seconds of interaction with the system). The learned control policy followed the reference trajectory for the end-effector with a mean error of 0.65 [mm] (standard deviation of 0.23 [mm]), and a maximum error of 1.08 [mm].

\section{MC-PILCO4PMS Experiments}\label{sec:experimentReal}
\begin{figure}[t]
\centering
  \includegraphics[width=0.9\linewidth]{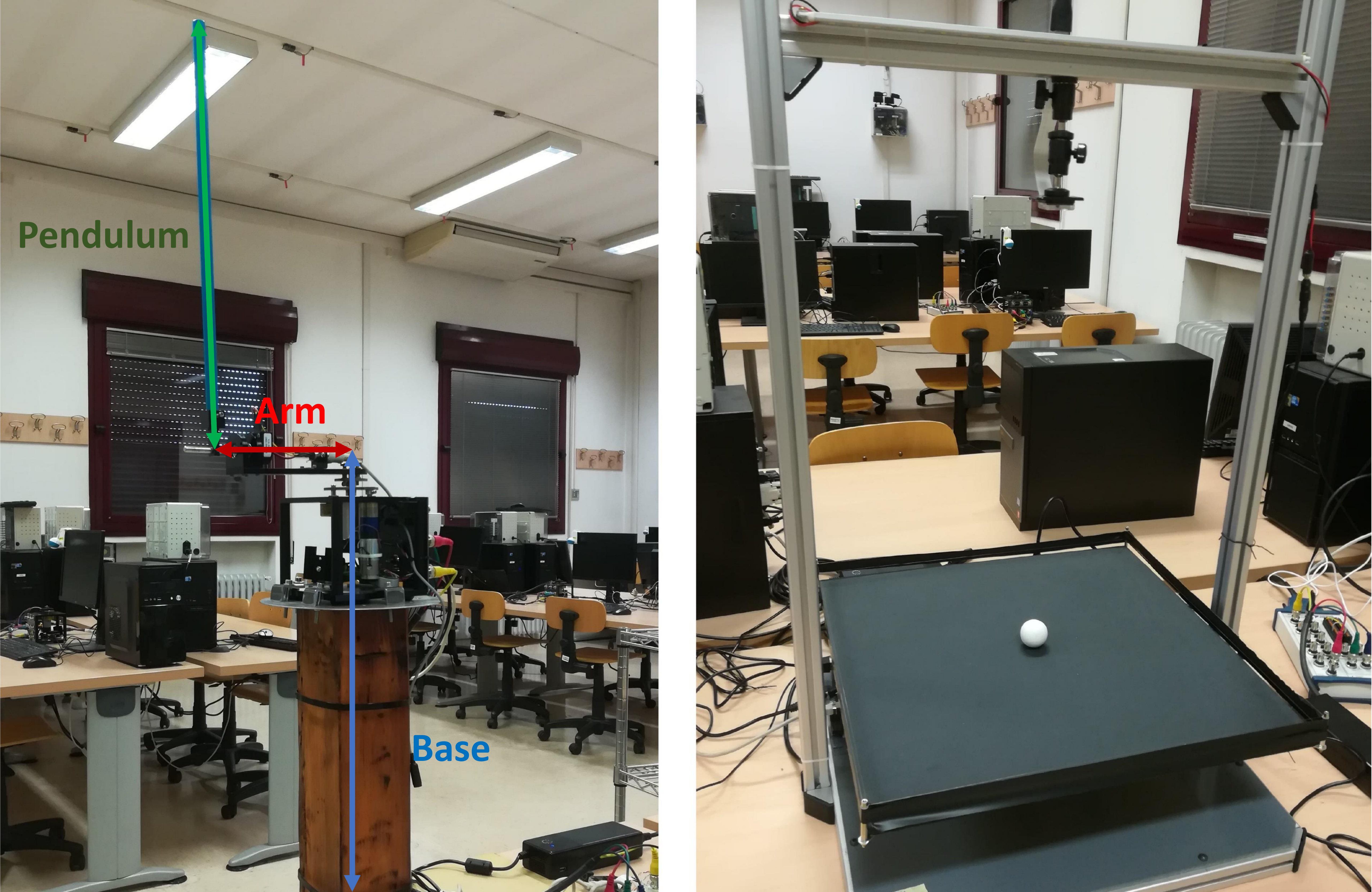}
  \caption{\small (Left) Furuta pendulum controlled in the upward equilibrium point by the learned policy. (Right) Ball-and-plate system.}
  \label{fig:real_systems}
\end{figure}

In this section, we provide the experimental results obtained by MC-PILCO4PMS. First, we propose a proof of concept on the simulated cart-pole benchmark, to better show the validity of of the concepts introduced in Section \ref{sec:MC-PILCO_RS}. Later, we we test MC-PILCO4PMS when applied to real systems. In particular, we experimented on two benchmark systems\footnote{A video of the experiments is available at \url{https://youtu.be/--73hmZYaHA}.}: a Furuta pendulum, and a ball-and-plate (Figure \ref{fig:real_systems}).
\subsection{MC-PILCO4PMS proof of concept}
\begin{figure}[t]
\centering
  \includegraphics[width=1\linewidth]{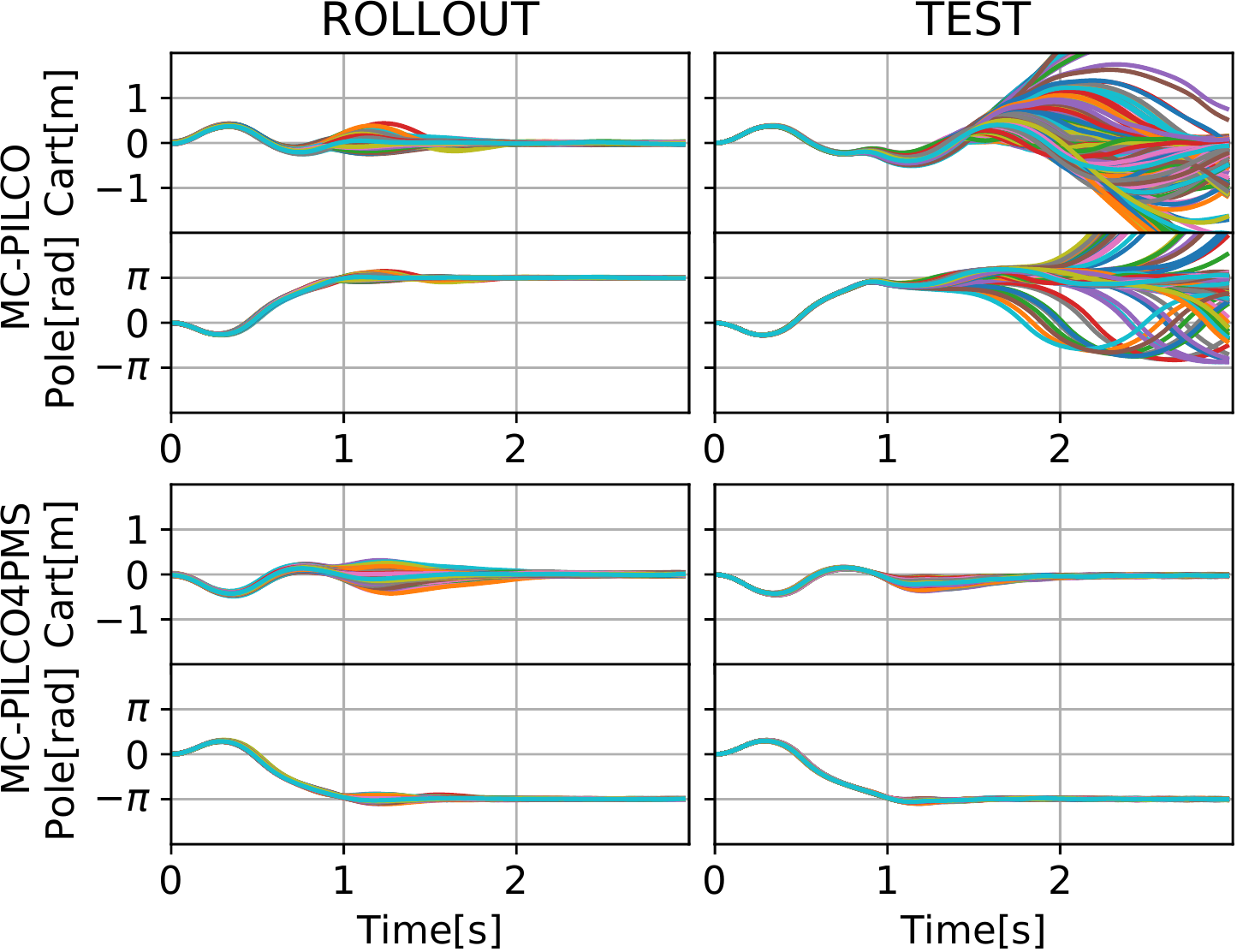}
  \caption{\small Comparison of 400 simulated particles rollout (left) and the trajectories performed applying repetitively the policy 400 times in the system (right) with the simulated cart-pole system. Each and all trajectories are shown with a line. Results obtained without simulating online filtering are on the top plots, while the ones obtained considering the low-pass filters are on the bottom. The plots refer to the policy learned after 5 trials with the system.}
  \label{fig:filt_comparison}
\end{figure}
Here, we test the relevance of modeling the presence of online estimators using the simulated cart-pole system, but adding assumptions that emulate a real world experiment. We considered the same physical parameters and the same initial conditions described in Section \ref{sec:ablation}, but assuming to measure only the cart position and the pole angle. We modeled a possible measurement system that we would have in the real world as an additive Gaussian i.i.d. noise with standard deviation $3\cdot 10^{-3}$. In order to obtain reliable estimates of the velocities, samples were collected at 30 [Hz]. The online estimates of the velocities were computed by means of causal numerical differentiation followed by a first order low-pass filter, with cutoff frequency 7.5 [Hz]. The velocities used to train the GPs were derived with the central difference formula. To verify the effectiveness of MC-PILCO4PMS (described in Section~\ref{sec:MC-PILCO_RS}) two policy functions were trained. The first policy is obtained with MC-PILCO by neglecting the presence of online filtering during policy optimization and assuming direct access to the state predicted by the model. On the contrary, the second policy is trained with MC-PILCO4PMS, which models the presence of the online estimators. Exploration data were collected with a random policy. To avoid dependencies on initial conditions, such as policy initialization and exploration data, we fixed the same random seed in both experiments. In Figure \ref{fig:filt_comparison}, we report the results of a Monte Carlo study with 400 runs. On the left, the final policy is applied to the learned models (ROLLOUT) and on the right to the cartpole system (TEST). 
Even though the two policies perform similarly when applied to the models, which is all can be tested offline, the results obtained by testing the policies in the cartpole system are significantly different. The policy optimized with modeling the presence of online filtering solves the task in all 400 attempts. In contrast, in several attempts, the first policy does not solve the task, due to delays and discrepancies introduced by the online filter and not considered during policy optimization. We believe that these considerations on how to manipulate the data during model learning and policy optimization might be beneficial for other algorithms than MC-PILCO.
\subsection{Furuta pendulum}
The Furuta pendulum (FP) \cite{cazzolato2011furuta} is a popular benchmark system used in nonlinear control and RL. The system is composed of two revolute joints and three links. The first link, called the base, is fixed and perpendicular to the ground. The second link, called arm, rotates parallel to the ground, while the rotation axis of the last link, the pendulum, is parallel to the principal axis of the second link, see Figure~\ref{fig:real_systems}. The FP is an under-actuated system as only the first joint is actuated. In particular, in the FP considered the horizontal joint is actuated by a DC servomotor, and the two angles are measured by optical encoders with 4096 [ppr]. The control variable is the motor voltage.
Let the state at time step $t$ be $\boldsymbol{x}_t = [\theta^h_t, \dot{\theta}^h_t, \theta^v_t, \dot{\theta}^v_t ]^T$, where $\theta^h_t$ is the angle of the horizontal joint and $\theta^v_t$ the angle of the vertical joint attached to the pendulum. The objective is to learn a controller able to swing-up the pendulum and stabilize it in the upwards equilibrium ($\theta_t^v=\pm\pi$ [rad]) with $\theta_t^h=0$ [rad]. The trial length is 3 seconds with a sampling frequency of 30 [Hz]. The cost function is defined as
\begin{equation}\label{eq:furuta cost}
    c(\boldsymbol{x}_t)=1-\text{exp}\left( -\left(\frac{\theta_t^h}{2}\right)^2  -\left(\frac{|\theta_t^v|-\pi}{2}\right)^2 \right) +c_{b}(\boldsymbol{x}_t),
\end{equation}
with
\begin{align*}
    c_{b}(\boldsymbol{x}_t) =& \frac{1}{1+\text{exp}\left(-10\left(-\frac{3}{4}\pi-\theta^h_t \right)\right)}\\
    &+\frac{1}{1+\text{exp}\left(-10\left(\theta^h_t-\frac{3}{4}\pi \right)\right)}\text{.}
\end{align*}

\begin{figure}[t]
\centering
  \includegraphics[width=0.91\linewidth]{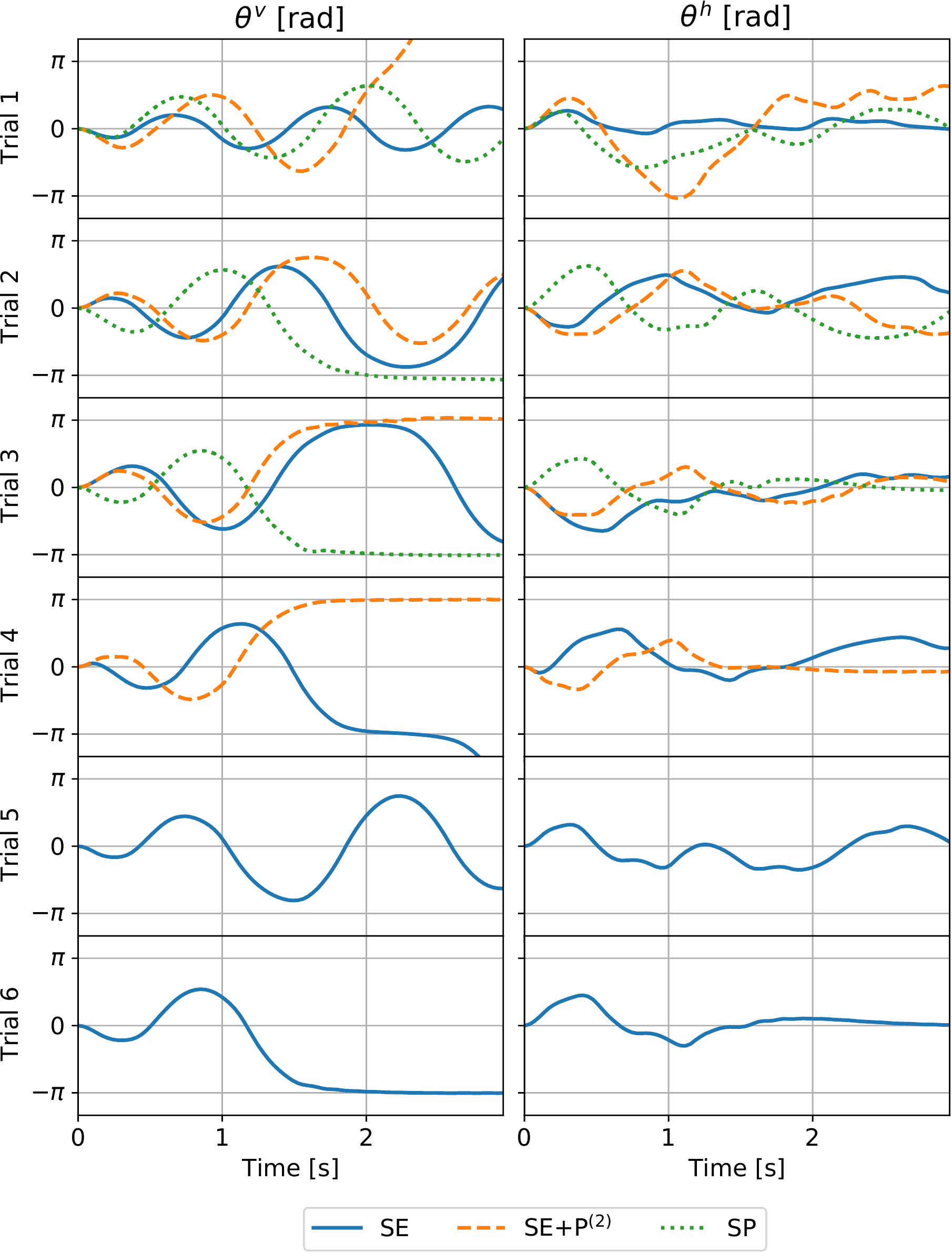}
  \caption{\small (Left) Pendulum angle's trajectories for each trial. (Right) Horizontal joint angle's trajectories for each trial. For all the kernels, the angles are plotted up to the trial that solved the task.}
  \label{fig:FP_results}
\end{figure}

The first part of the function in \eqref{eq:furuta cost} aims at driving the two angles towards $\theta_t^h=0$ and $\theta_t^v=\pm\pi$, while $c_{b}(\boldsymbol{x}_t)$ penalizes solutions where $\theta_t^h \leq -\frac{3}{4} \pi$ or  $\theta_t^h \geq \frac{3}{4} \pi$. We set those boundaries to avoid the risk of damaging the system if the horizontal joint rotates too much. Offline estimates of velocities for the GP model have been computed by means of central differences. For the online estimation, we used causal numerical differentiation: $\dot{\boldsymbol{q}}_t = (\boldsymbol{q}_{t}-\boldsymbol{q}_{t-1})/(T_s)$, where $T_s$ is the sampling time. Instead of $\boldsymbol{x}_t$, we considered the extended state $\boldsymbol{x}^*_t=[\dot{\theta}^h_t, \dot{\theta}^v_t, sin(\theta^h_t),cos(\theta^h_t),sin(\theta^v_t),cos(\theta^v_t)]^T$ in GP input. The policy is a \textit{squashed-RBF-network} with $n_b=200$ basis functions that receives as input $[(\theta^h_t-\theta^h_{t-1})/{T_s}, (\theta^v_t-\theta^v_{t-1})/T_s, sin(\theta^h_t),cos(\theta^h_t),sin(\theta^v_t),cos(\theta^v_t)]^T$. The exploration trajectory has been obtained using as input a sum of ten sine waves of random frequencies and same amplitudes. The initial state distribution is assumed to be $\mathcal{N}([0,0,0,0]^T,\text{diag}([5 \cdot 10^{-3},5 \cdot 10^{-3},5 \cdot 10^{-3},5 \cdot 10^{-3}])$. The GP reduction thresholds were set to $10^{-3}$. We solved the task using the three different choices of kernel functions described in Section \ref{sec:kernels}: squared exponential (SE), squared exponential + polynomial of degree $d$ (SE+$\text{P}^{(d)}$) and semi-parametrical (SP)\footnote{SP basis functions can be obtained by isolating, in each ODE defining FP laws of motion, all the linearly related state-dependent components. In particular, we have $
\phi_{\dot{\theta}^h} (\boldsymbol{x}, u) = [(\dot{\theta}^v)^2 sin(\theta^v), \dot{\theta}^h \dot{\theta}^v sin(2\theta^v), \dot{\theta}^h, u] $
 for the arm velocity GP, and $
\phi_{\dot{\theta}^v} (\boldsymbol{x}, u) = [(\dot{\theta}^h)^2 sin(2\theta^v), \dot{\theta}^v, sin(\theta^v), u\;cos(\theta^v)] $ for the pendulum velocity GP.}. In Figure \ref{fig:FP_results}, we show the resulting trajectories for each trial.
MC-PILCO4PMS managed to learn how to swing up the Furuta pendulum in all cases. It succeeded at trial 6 with kernel SE, at trial 4 with kernel SE+$\text{P}^{(2)}$, and at trial 3 with SP kernel. These experimental results confirm the higher data efficiency of more structured kernels and the advantage of allowing any kernel function offered by our MBRL method. Moreover, we can observe the effectiveness of the cost function \eqref{eq:furuta cost} in keeping $\theta_t^h$ always inside the desired boundaries in all the trials and for any kernel tested. Considering penalties similar to $c_b(\boldsymbol{x}_t)$ inside the cost function could be enough to handle soft constraints also in other scenarios.

\subsection{Ball-and-plate}
The ball-and-plate system is composed of a square plate that can be tilted in two orthogonal directions by means of two motors. On top of it, there is a camera to track the ball and measure its position on the plate. Let $(b^x_t,b^y_t)$ be the position of the center of the ball along X-axis and Y-axis, while $\theta^{(1)}_t$ and $\theta^{(2)}_t$ are the angles of the two motors tilting the plate, at time $t$. So, the state of the system is defined as $\boldsymbol{x}_t = [b^x_t,b^y_t,\dot{b}^x_t,\dot{b}^y_t,\theta^{(1)}_t,\theta^{(2)}_t,\dot{\theta}^{(1)}_t,\dot{\theta}^{(2)}_t]^T$. The drivers of the motors allow only position control, and do not provide feedback about the motors angles. To keep track of the motor angles, we defined the control actions as the difference between two consecutive reference values sent to the motor controllers, and we limited the maximum input to a sufficiently small value, such that the motor controllers are able to reach the target angle within the sampling time. Then, in first approximation, the reference angles and the motor angles coincide, and we have $u_t^{(1)} = \theta^{(1)}_{t+1}-\theta^{(1)}_t$ and $u_t^{(2)} = \theta^{(2)}_{t+1}-\theta^{(2)}_t$.
%
The objective of the experiment is to learn how to control the motor angles in order to stabilize the ball around the center of the plate. Notice that the control task, with the given definition of inputs, is particularly difficult because the policy must learn to act in advance, and not only react to changes in the ball position. The cost function is defined as
\begin{equation*}
    c(\boldsymbol{x}_t)=1-\text{exp}\left(-g_t(\boldsymbol{x}_t) \right), \qquad \text{with}
\end{equation*}
\begin{equation*}
    g_t(\boldsymbol{x}_t) = \left(\frac{b^x_t}{0.15}\right)^2 +\left(\frac{b^y_t}{0.15}\right)^2  +\left(\theta_t^{(1)}\right)^2  +\left(\theta_t^{(2)}\right)^2.
\end{equation*}
The trial length is 3 seconds, with a sampling frequency of 30 [Hz]. Measurements provided by the camera are very noisy, and cannot be used directly to estimate velocities from positions. We used a Kalman smoother for the offline filtering of ball positions ($b^x_t, b^y_t$) and associated velocities ($\dot{b}^x_t, \dot{b}^y_t$). In the control loop, instead, we used a Kalman filter \cite{kalman1960new} to estimate online the ball state from noisy measures of positions. Concerning the model, we need to learn only two GPs predicting the evolution of the ball velocity because we directly control motor angles, hence, their evolution is assumed deterministic. GP inputs, $\tilde{\boldsymbol{x}}_t = [\boldsymbol{x}^*_t, u_t]$, include an extended version of the state,  $\boldsymbol{x}^*_t=[b^x_t,b^y_t,\dot{b}^x_t,\dot{b}^y_t,sin(\theta^{(1)}_t),cos(\theta^{(1)}_t),sin(\theta^{(2)}_t),cos(\theta^{(2)}_t),(\theta^{(1)}_t-\theta^{(1)}_{t-1})/T_s,(\theta^{(2)}_t-\theta^{(2)}_{t-1})/T_s]^T$ where angles have been replaced by their sines and cosines, and motor angular velocities have been estimated with causal numerical differentiation ($T_s$ is the sampling time). The SE+$\text{P}^{(1)}$ kernel \eqref{eq:SE+Pkernel} is used, where the linear kernel acts only on a subset of the model inputs, $\tilde{\boldsymbol{x}}^{lin}_t=[sin(\theta^{(1)}_t), sin(\theta^{(2)}_t), cos( \theta^{(1)}_t), cos(\theta^{(2)}_t), u_t]$. We diminished the GP reduction threshold to $10^{-4}$ w.r.t. the FP experiment because of the small distances the ball can cover in a time step. The policy is a multi-output RBF network \eqref{eq:policy}, with $n_b=400$ basis functions, that receives as inputs the estimates of  $(b^x_t,b^y_t,\dot{b}^x_t,\dot{b}^y_t,\theta^{(1)}_t, \theta^{(1)}_{t-1},\theta^{(2)}_t,\theta^{(2)}_{t-1})$ computed with the Kalman filter; maximum angle displacement is $u_{max}=4$ [deg] for both motors. The policy optimization parameters used were the same described in Table \ref{tab: standard_opt_setup}, with the difference that we set $\alpha_{lr}=0.006$ as initial learning rate. The reduction of the learning rate is related to the use of small length-scales in the cost function, that are necessary to cope with the small range of movement of the ball. For the same reason, we set also $\alpha_{lr_{min}}=0.0015$ and $\sigma_s=0.05$. Initial exploration is given by two different trials, in which the control signals are two triangular waves perturbed by white noise. Mostly during exploration and initial trials, the ball might touch the borders of the plate. In those cases, we kept data up to the collision instant. A peculiarity of this experiment in comparison to the others seen before is a wide range of initial conditions. In fact, the ball could be positioned anywhere on the plate's surface, and the policy must control it to the center. The initial distribution of $b^x_0$ and $b^y_0$ is a uniform $\mathcal{U}(-0.15,0.15)$, which covers almost the entire surface (the plate is a square with sides of about 0.20 [m]). For the other state components, $\theta^{(1)}_t$ and $\theta^{(2)}_t$, we assumed tighter initial distributions $\mathcal{U}(-10^{-6},10^{-6})$.
MC-PILCO4PMS managed to learn a policy able to control the ball around the center starting from any initial position after the third trial, 11.33 seconds of interaction with the system.
\begin{figure}[t]
\centering
  \includegraphics[width=0.87\linewidth]{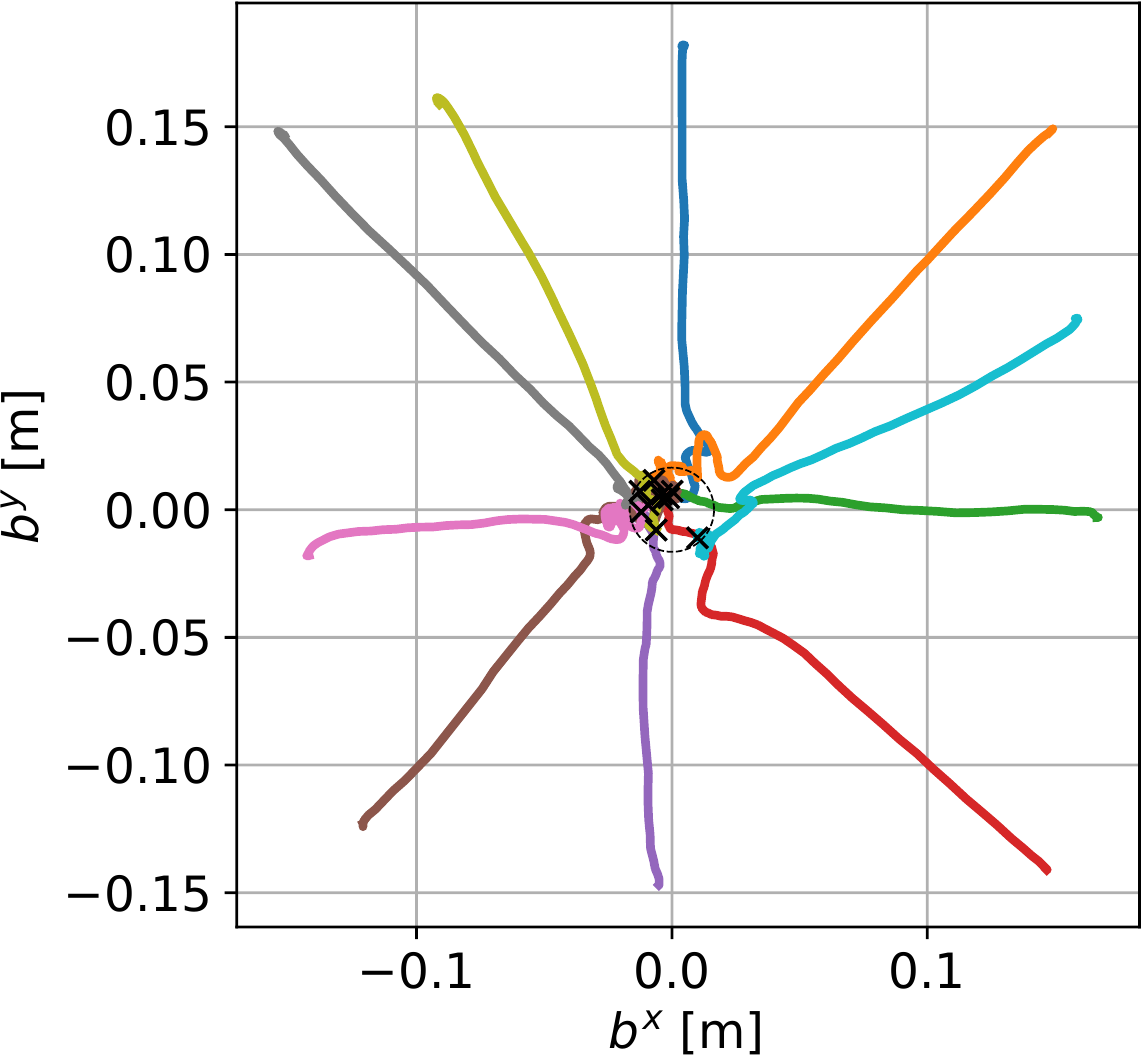}
  \caption{\small Ten different ball trajectories obtained under the final policy learned by MC-PILCO4PMS. Steady-state positions are marked with black crosses. The dashed circle has the same diameter as the ball.}
  \label{fig:b&p}
\end{figure}
We tested the learned policy starting from ten different points, see Figure \ref{fig:b&p}. The mean steady-state error, i.e., the average distance of the final ball position from the center observed in the ten trials, was 0.0099 [m], while the maximum measured error was 0.0149 [m], which is lower than the ball radius of 0.016 [m].

\section{Conclusions}\label{sec:conclusions}
In this paper, we have presented the MBRL algorithm MC-PILCO. The proposed framework uses GPs to derive a probabilistic model of the system dynamics, and updates the policy parameters through a gradient-based optimization that exploits the \emph{reparameterization trick} and approximates the expected cumulative cost relying on a Monte Carlo approach. Compared to similar algorithms proposed in the past, our Monte Carlo approach worked by focusing on two aspects, that are (i) proper selection of the cost function, and (ii) introduction of dropout during policy optimization. Extensive experiments on the simulated cart-pole benchmark confirm the effectiveness of the proposed solution, and show the relevance of the two aforementioned aspects when optimizing the policy combining the \emph{reparameterization trick} with particle-based methods. Particles-based approximation offers other two advantages in comparison to the moment-matching approach of PILCO, namely, the possibility of using structured kernels, such as polynomial kernels and semi-parametrical kernels, and the ability of handling multimodal distributions. In particular, experimental results show that the use of structured kernels can increase data efficiency, reducing the interaction-time required to learn the task. MC-PILCO was also used to learn from scratch a joint-space controller for a (simulated) robotic manipulator, proving able to handle such a relatively high-DoF task. Moreover, we compared MC-PILCO with PILCO and Black-DROPS (two state-of-the-art GP-based MBRL algorithms) on the cart-pole benchmark. MC-PILCO outperformed both algorithms in this scenario, exhibiting better data efficiency and asymptotic performance.

Furthermore, we analyzed common problems that arise when trying to apply MBRL to real systems. In particular, we focused on systems with partially measurable states (e.g., mechanical systems) which are particularly relevant in real applications. In this context, we proposed a modified version of our algorithm, called MC-PILCO4PMS, through which we verified the importance of taking into account the state estimators used in the real system during policy optimization. Results have been validated on two different real setups, specifically, a Furuta pendulum and a ball-and-plate system.

In future works, we are interested in testing the proposed algorithms in more challenging scenarios, e.g., manipulation tasks in real world environments. The issues regarding the impossibility of measuring directly the velocity states tackled in MC-PILCO4PMS could be further analyzed by considering the recently introduced "Velocity-free" framework \cite{derivative_free_RL}. Finally, the application to manipulation tasks will also require the introduction of safe exploration techniques and guarantees from the state-of-the-art in safe RL \cite{garcia2015safeRL}.



%

\bibliographystyle{unsrt}
\bibliography{references}

\vfill
\textcopyright 2022 IEEE. Personal use of this material is permitted. Permission from IEEE must be obtained for all other uses, in any current or future media, including reprinting/republishing this material for advertising or promotional purposes, creating new collective works, for resale or redistribution to servers or lists, or reuse of any copyrighted component of this work in other works.

\end{document}